\documentclass{article}
\usepackage{arxiv}
\usepackage[T1]{fontenc}
\usepackage[utf8]{inputenc}
\usepackage{amsmath}
\usepackage{amssymb}
\usepackage{multirow}
\usepackage{graphicx}
\usepackage{subcaption}
\usepackage{enumitem}
\usepackage{url}
\usepackage{algorithm}
\usepackage[noend]{algpseudocode}
\usepackage{booktabs}
\usepackage{natbib}
\usepackage{hyperref}
\setlist[enumerate]{noitemsep, topsep=0pt}
\setlist[itemize]{noitemsep, topsep=0pt}

\newcommand{\myvec}[1]{\boldsymbol{#1}}

\newcommand{\mysp}[1]{\mathtt{#1}}
\newcommand{\mysim}{\cdot}
\newcommand{\mybundle}{+}
\newcommand{\mybind}{\circledast}
\newcommand{\myinvert}[1]{#1^{-1}}
\newcommand{\spspace}{\mathbb{R}^{N}}
\newcommand{\featurespace}{\mathbb{R}^D}
\newcommand{\gridspace}{\mathbb{G}}
\newcommand{\pixelspace}{\mathbb{P}}
\newcommand{\myoperation}[1]{\textsc{#1}}

\title{Vector Symbolic Algebras for the Abstraction and Reasoning Corpus}

\author{
    Isaac Joffe \\
    Centre for Theoretical Neuroscience \\
	David R. Cheriton School of Computer Science \\
	University of Waterloo \\
	Waterloo, ON \\
	\texttt{ijoffe@uwaterloo.ca} \\
	\And
    Chris Eliasmith \\
    Centre for Theoretical Neuroscience \\
	Department of Philosophy \\
	Department of Systems Design Engineering \\
	University of Waterloo \\
	Waterloo, ON \\
	\texttt{celiasmith@uwaterloo.ca} \\
}

\date{}

\begin{document}

\maketitle

\begin{abstract}
    The Abstraction and Reasoning Corpus for Artificial General Intelligence (ARC-AGI) is a generative, few-shot fluid intelligence benchmark. Although humans effortlessly solve ARC-AGI, it remains extremely difficult for even the most advanced artificial intelligence systems. Inspired by methods for modelling human intelligence spanning neuroscience to psychology, we propose a cognitively plausible ARC-AGI solver. Our solver integrates System 1 intuitions with System 2 reasoning in an efficient and interpretable process using neurosymbolic methods based on Vector Symbolic Algebras (VSAs). Our solver works by object-centric program synthesis, leveraging VSAs to represent abstract objects, guide solution search, and enable sample-efficient neural learning. Preliminary results indicate success, with our solver scoring $10.8\%$ on ARC-AGI-1-Train and $3.0\%$ on ARC-AGI-1-Eval. Additionally, our solver performs well on simpler benchmarks, scoring $94.5\%$ on Sort-of-ARC and $83.1\%$ on 1D-ARC---the latter outperforming GPT-4 at a tiny fraction of the computational cost. Importantly, our approach is unique; we believe we are the first to apply VSAs to ARC-AGI and have developed the most cognitively plausible ARC-AGI solver yet. Our code is available at: \url{https://github.com/ijoffe/ARC-VSA-2025}.
\end{abstract}

\section{Introduction}
\label{sec:introduction}

The Abstraction and Reasoning Corpus for Artificial General Intelligence \citep[ARC-AGI;][]{chollet2019measureintelligence,arcagi2} challenges the tremendous progress deep learning has made in artificial intelligence (AI). ARC-AGI is a fluid intelligence benchmark comprising a collection of grid prediction tasks (see Figs. \ref{fig:task_0962bcdd} and \ref{fig:task_54d9e175}). The goal of the test-taker, human or AI, is simple: given a few pairs of input and output grids containing abstract symbols, determine the rules underlying the symbol transformations and use this understanding to predict the output grids corresponding to lone test input grids. ARC-AGI is easy for humans. Human performance is estimated to be $85.0\%$ and all tasks were solved by at least one of two experts \citep{arcprize2024}, indicating human intelligence can completely solve the benchmark. Conversely, ARC-AGI is hard for AI. Despite ample focus and investment over multiple $\$1{,}000{,}000$ competitions \citep{kaggleprize2024,kaggleprize2025}, the benchmark remains unbeaten. ARC-AGI has proved resistant to old and new deep learning techniques, including the large language model (LLM)---OpenAI's original GPT-3 \citep{gpt3} scored $0.0\%$ \citep{arcprize2024} notwithstanding claims of emergent reasoning \citep{emergent_llm}. The chasm between human and AI performance on ARC-AGI suggests fundamental problems with leading approaches to AI.

Two such problems often identified in the literature are sample-efficient learning and explicit reasoning \citep{turing_dl,schmidhuber_bind,hinton_dl}. Deep learning's success requires vast amounts of often-labelled high-quality training data, making it ineffective in the sample-few regime. Additionally, deep learning systems struggle with representing explicit rules, sometimes manifesting in poor out-of-distribution generalization. Vector Symbolic Algebras (VSAs), a neurosymbolic AI method introducing syntactic structure into high-dimensional distributed representations, hold promise to overcome these two limitations that ARC-AGI targets. VSAs specify high-dimensional vectors to represent structured low-dimensional data, discrete or continuous. Thus, instead of learning useful embeddings from huge datasets, a VSA can be used to encode data with desired inductive biases. VSAs also define operators for the composition and systematic processing of data. Thus, instead of learning shortcut transformations that may generalize poorly to unseen inputs, a VSA, despite relying on distributed representations, can be used to express certain operations analytically. Additionally, VSAs support discrete search and neural learning, both of which are helpful for ARC-AGI. These strengths of VSAs, along with their ability to effectively model cognitive processes in a biologically plausible manner \citep{eliasmith2003neural,eliasmith2013how}, make them an excellent candidate for helping solve ARC-AGI.

We propose a novel, neurosymbolic, cognitively plausible ARC-AGI solver. Our solver uses VSAs, a general cognitive modelling framework, to bridge intuition and reason and capture the efficiency and interpretability of human intelligence. We believe we are the first to apply VSAs to ARC-AGI and, in doing so, have developed the most cognitively plausible ARC-AGI solver yet. The rest of this paper is organized as follows:
\begin{itemize}
    \item Section \ref{sec:preliminaries} provides the background knowledge necessary to understand our work. We introduce the VSA used, discuss motivating history, and describe ARC-AGI further.
    \item Section \ref{sec:methods} explains our approach. We detail and justify each part of our solver's design.
    \item Section \ref{sec:results} presents our results. We examine our solver's performance on ARC-AGI and related datasets.
    \item Section \ref{sec:discussion} analyzes our approach. We outline our solver's strengths and report its weaknesses.
    \item Section \ref{sec:conclusion} summarizes our work.
\end{itemize}

\section{Preliminaries}
\label{sec:preliminaries}

\subsection{Vector Symbolic Algebras}

Before explaining how our solver---which uses VSAs extensively---works, we describe VSAs further. VSAs, also known as hyperdimensional computing, are a family of algebras defining operations over a high-dimensional vector space. In this work, we use Holographic Reduced Representations \citep[HRRs;][]{plate1991,plate1995,plate2003}, one particular VSA.

\subsubsection{Holographic Reduced Representations}

HRRs define four operations: \textit{similarity}, \textit{binding}, \textit{bundling}, and \textit{inversion}. Following \citet{furlong2022fractional}, we describe each.

Similarity, denoted by the $\mysim : \spspace \times \spspace \rightarrow \mathbb{R}$ operator, operates on two vectors, $\myvec{a}$ and $\myvec{b}$, to produce a scalar, $s = \myvec{a} \mysim \myvec{b}$. The resultant value represents the semantic similarity between the vectors. Similarity is implemented as the vector dot product: $s = \myvec{a} \mysim \myvec{b} = \langle \myvec{a}, \myvec{b} \rangle = \left\lVert \myvec{a} \right\rVert \left\lVert \myvec{b} \right\rVert \cos{\theta_{\myvec{ab}}}$. Because we usually work with unit-length vectors (i.e., $\left\lVert \myvec{a} \right\rVert = \left\lVert \myvec{b} \right\rVert = 1$), the dot product is equivalent to cosine similarity and, thus, $s$ falls between $-1$ and $1$. Similarity is commutative. In high-dimensional vector spaces, the similarity between any two randomly-generated vectors is likely to be extremely small. Similarity can be used to \textit{clean-up} noisy vectors by replacing the noisy vector with the one from a library of known vectors to which the noisy vector is most similar.

Bundling, denoted by the $\mybundle : \spspace \times \spspace \rightarrow \spspace$ operator, combines two vectors, $\myvec{a}$ and $\myvec{b}$, into one, $\myvec{c} = \myvec{a} \mybundle \myvec{b}$. The resultant vector is similar to both input vectors (i.e., $\myvec{c} \cdot \myvec{a} \gg 0$ and $\myvec{c} \cdot \myvec{b} \gg 0$); bundling gathers multiple vectors into a single representation, producing a superposition of those vectors. Bundling is implemented as element-wise vector addition. Bundling is commutative and associative.

Binding, denoted by the $\mybind : \spspace \times \spspace \rightarrow \spspace$ operator, combines two vectors, $\myvec{a}$ and $\myvec{b}$, into one, $\myvec{c} = \myvec{a} \mybind \myvec{b}$. The resultant vector is dissimilar to both input vectors (i.e., $\myvec{c} \cdot \myvec{a} \simeq 0$ and $\myvec{c} \cdot \myvec{b} \simeq 0$); binding associates two vectors, producing a new vector unlike any existing vectors.  Binding is implemented as circular convolution: $\myvec{c} = \myvec{a} \mybind \myvec{b} = \mathcal{F}^{-1} \left\{ \mathcal{F} \left\{ \myvec{a} \right\} \odot \mathcal{F} \left\{ \myvec{b} \right\} \right\}$, where $\mathcal{F}$ is the discrete Fourier transform (DFT) and $\odot$ denotes Hadamard, or element-wise, multiplication. Binding is commutative and associative.

Inversion, denoted by the $\myinvert{} : \spspace \rightarrow \spspace$ operator, converts one vector, $\myvec{a}$, into another, $\myinvert{\myvec{a}}$. The resultant vector bound with the input vector approximately produces the binding identity vector (i.e., $\myvec{a} \mybind \myinvert{\myvec{a}} \simeq \mathbf{I}$ where $\myvec{a} \mybind \mathbf{I} = \myvec{a}$). Inversion can be used to \textit{unbind} bound vectors: given a bound vector, $\myvec{c} = \myvec{a} \mybind \myvec{b}$, and one of its constituent vectors, $\myvec{a}$, the other can be extracted by binding with the inverse of the known constituent (i.e., $\myvec{c} \mybind \myinvert{\myvec{a}} = (\myvec{a} \mybind \myvec{b}) \mybind \myinvert{\myvec{a}} \simeq \myvec{b}$).

\begin{figure}
    \centering
    \begin{minipage}{0.47\textwidth}
        \centering
        \includegraphics[width=\linewidth]{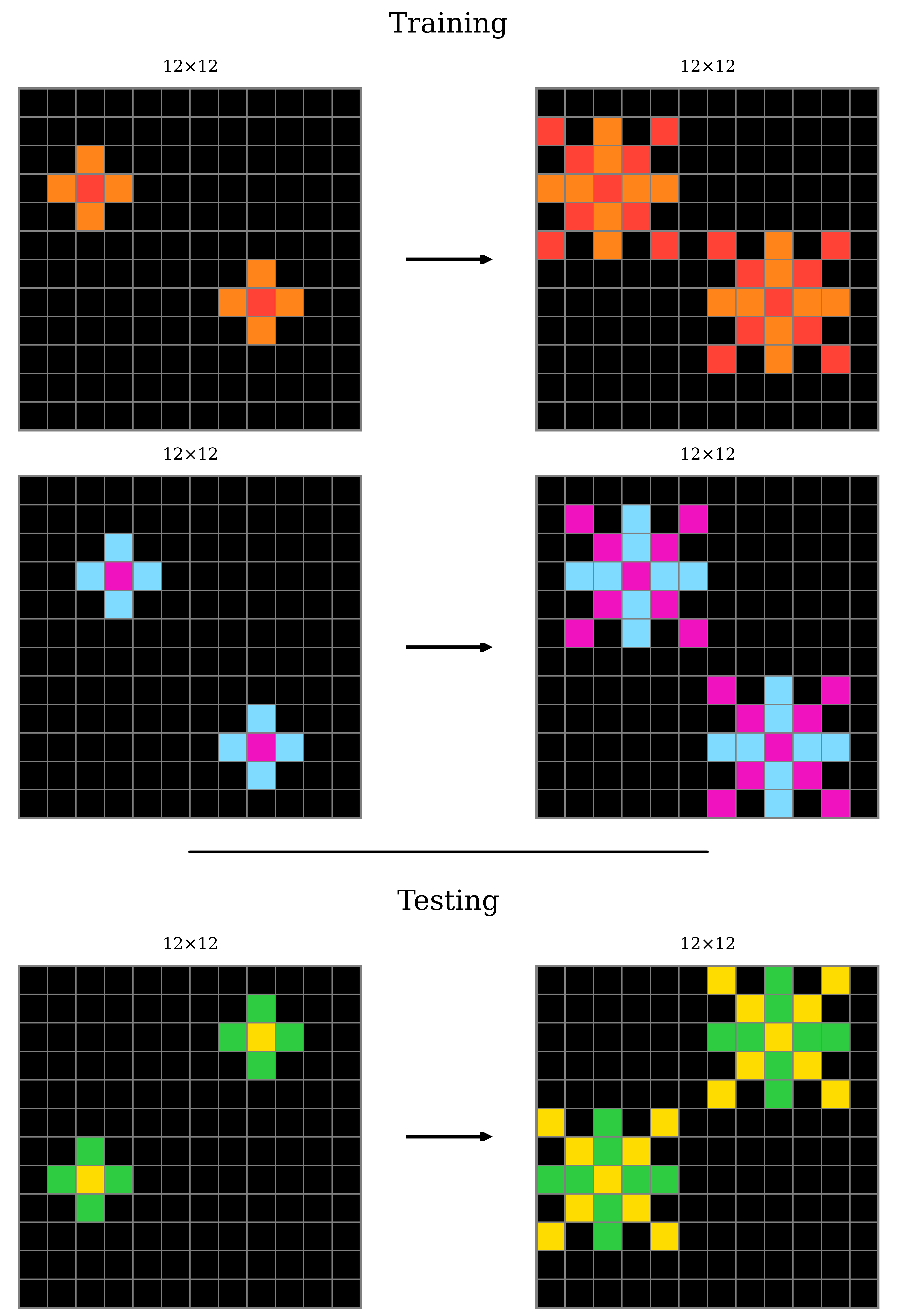}
        \caption{ARC-AGI task \texttt{0962bcdd}. Here, the implicit pattern is to transform each multi-coloured object into a larger object.}
        \label{fig:task_0962bcdd}
    \end{minipage}
    \hfill
    \begin{minipage}{0.47\textwidth}
        \centering
        \includegraphics[width=\linewidth]{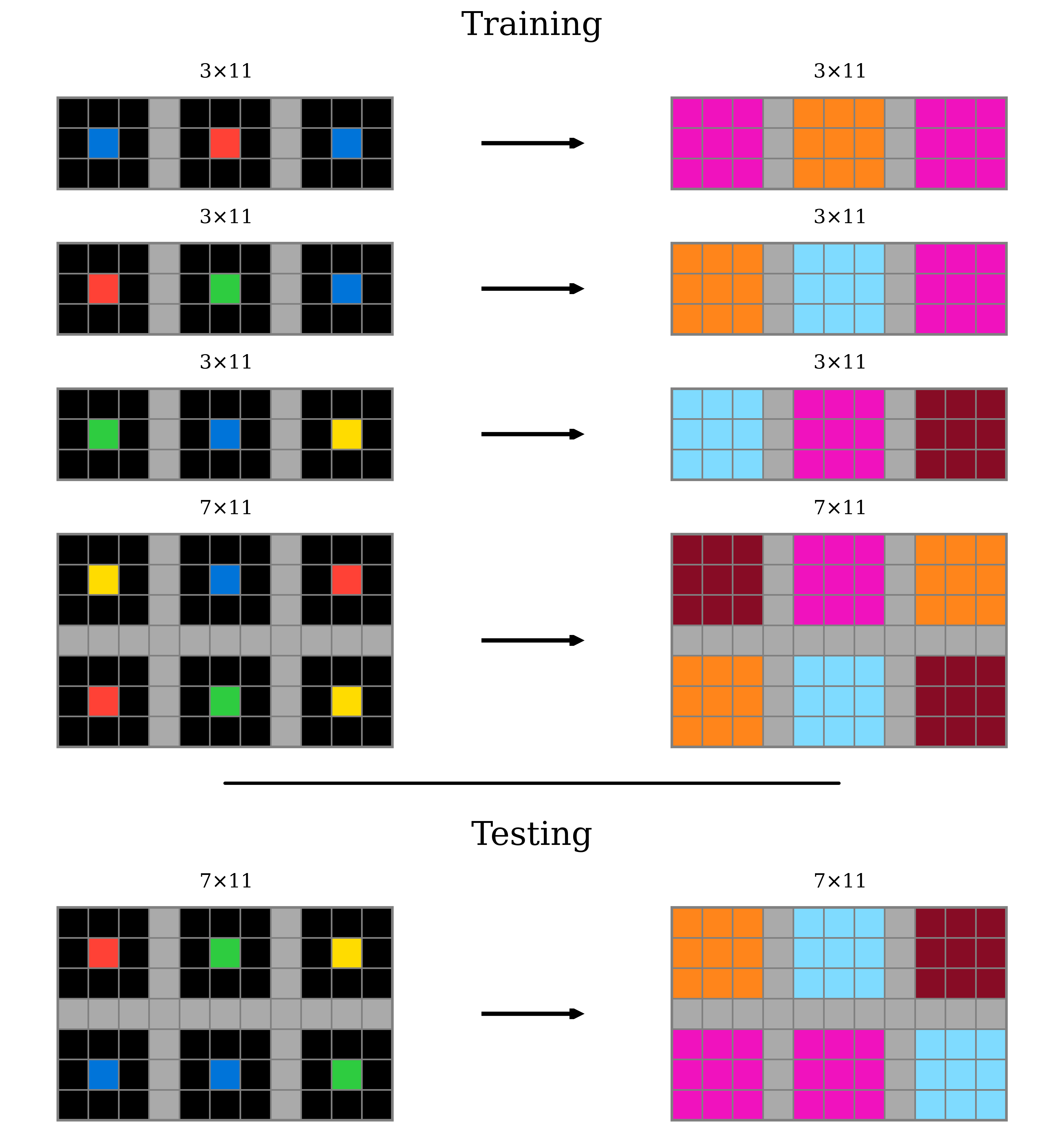}
        \caption{ARC-AGI task \texttt{54d9e175}. Here, the implicit pattern is to colour each black region partitioned by the grey pixels according to the colour of the lone pixel at its centre (blue to pink, red to orange, green to cyan, and yellow to purple).}
        \label{fig:task_54d9e175}
    \end{minipage}
\end{figure}

These four operations can be combined to encode and compute useful functions on complex data structures, such as lists \citep{choo2010} and graphs \citep{laube2024}. For example, we can use a \textit{slot-filler} representation to store multiple features of an item in a compressed but interpretable format. Given known slot vectors $\mysp{SIZE}, \mysp{SPECIES} \in \spspace$, we can encode a small dog as $\myvec{a} = \mysp{SIZE} \mybind \mysp{SMALL} + \mysp{SPECIES} \mybind \mysp{DOG}$ and a large cat as $\myvec{b} = \mysp{SIZE} \mybind \mysp{LARGE} + \mysp{SPECIES} \mybind \mysp{CAT}$, for instance. We can then query information about each object by unbinding: $\myvec{a} \mybind \myinvert{\mysp{SIZE}} \simeq \mysp{SMALL}$ and $\myvec{b} \mybind \myinvert{\mysp{SPECIES}} \simeq \mysp{CAT}$. The former query becomes
\begin{equation}
    \begin{aligned}
        \myvec{a} \mybind \myinvert{\mysp{SIZE}} & = (\mysp{SIZE} \mybind \mysp{SMALL} + \mysp{SPECIES} \mybind \mysp{DOG}) \mybind \myinvert{\mysp{SIZE}} \\
        & = \mysp{SIZE} \mybind \mysp{SMALL} \mybind \myinvert{\mysp{SIZE}} + \mysp{SPECIES} \mybind \mysp{DOG} \mybind \myinvert{\mysp{SIZE}}  \\
        & \simeq \mysp{SMALL} + \mysp{SPECIES} \mybind \mysp{DOG} \mybind \myinvert{\mysp{SIZE}} \\
        & \simeq \mysp{SMALL}
    \end{aligned}
\end{equation}
which works because $\mysp{SPECIES} \mybind \mysp{DOG} \mybind \myinvert{\mysp{SIZE}}$ behaves as noise that can be eliminated by clean-up.

Furthermore, we can encode a list of $k$ items as $\myvec{m} = \sum_{i}^{k}{\mysp{ITEM}_i \mybind \mysp{ONE}^i}$, where $\mysp{ONE}^i$ is the $i$-th index of the sequence \citep{choo2010}. The slots, $\mysp{ONE}^i$, are defined according to a recursively-bound index vector, $\mysp{ONE}$, as
\begin{equation}
    \begin{aligned}
        \mysp{TWO} & = \mysp{ONE} \mybind \mysp{ONE} = \mysp{ONE}^2 \\
        \mysp{THREE} & = \mysp{TWO} \mybind \mysp{ONE} = \mysp{ONE} \mybind \mysp{ONE} \mybind \mysp{ONE} = \mysp{ONE}^3 \\
        \mysp{FOUR} & = \ldots
    \end{aligned}
\end{equation}
where use of the exponentiation symbol is purely for notational convenience; the underlying operation is not repeated multiplication, but repeated binding. With this representation, the index of a known element in the list or the element at a known index in the list can be queried.

The Semantic Pointer Architecture \citep[SPA;][]{eliasmith2013how} is an architecture for modelling biological cognition using HRRs. The SPA has been used \citep{rasmussen2010a,rasmussen2011} to solve the Raven's Progressive Matrices \citep[RPMs;][]{raven1936} by which ARC-AGI was inspired. This RPM solver worked by generating VSA representations of each cell in the grid based on perceptual outputs, unbinding each cell's representation from its subsequent cell's representation to obtain transformation vectors, averaging these transformation vectors to obtain a single rule vector, and applying this rule vector to the second-last cell to predict the final cell. ARC-AGI is a significantly harder problem because of its less constrained format (e.g., the requirement to generate solutions from scratch) and content (e.g., each task is distinct), but this past success suggests VSAs hold potential.

\subsubsection{Spatial Semantic Pointers}

Typical HRRs can represent and express a multitude of discrete, but not continuous, information structures and operations. Expanding on early suggestions \citep{plate1992}, Spatial Semantic Pointers \citep[SSPs;][]{komer2019,komer2020} extend HRRs to continuous spaces. These representations can model \citep{dumont2020gridcell} grid cell neurons observed in the brain \citep{humangridcells2010,ratgridcells2005} and can be used to solve challenging spatial reasoning tasks \citep{dumont2022pathint,dumont2023slam,dumont2025}.

An SSP encoding defines a mapping $\phi : \featurespace \rightarrow \spspace$ from a low, $D$-dimensional vector space, \textit{feature space}, to a high, $N$-dimensional vector space, \textit{SSP space}, based on a generalization of recursive binding into fractional binding. To understand this, consider a one-dimensional feature space. Before, the vector representing the $n$-th index of the sequence, $\mysp{ONE}^n$, was computed by binding $\mysp{ONE}$ with itself $n$ times. However, we can also compute $\mysp{ONE}^n$ as
\begin{equation}
    \begin{aligned}
        \mysp{ONE}^n & = \underbrace{\mysp{ONE} \mybind \ldots \mybind \mysp{ONE}}_{n \text{ times}} \\
        & = \mathcal{F}^{-1} \left\{ \underbrace{\mathcal{F} \left\{ \mysp{ONE} \right\} \odot \ldots \odot \mathcal{F} \left\{ \mysp{ONE} \right\}}_{n \text{ times}} \right\} \\
        & = \mathcal{F}^{-1} \left\{ \mathcal{F} \left\{ \mysp{ONE} \right\} ^n \right\} \\
    \end{aligned}
\end{equation}
where $^n$ represents element-wise exponentiation. This formulation allows natural generalization from $n \in \mathbb{Z}^{+}$ to $x \in \mathbb{R}$; instead of encoding discrete values by repeatedly binding the vector with itself, we can encode discrete or continuous values by directly exponentiating the DFT of the vector and taking the inverse DFT of this result. This formulation can be further generalized to represent multi-dimensional features, $\myvec{x} \in \featurespace$, by binding together the vectors representing each feature dimension. For example, consider the two-dimensional feature space $\mathbb{R}^2$. Given basis or axis vectors $\mysp{X}, \mysp{Y} \in \spspace$, an arbitrary point in the feature space $\myvec{x} = (x, y)$ can be expressed in the SSP space as $\phi(\myvec{x}) = \mysp{X}^{x} \mybind \mysp{Y}^{y} = \mathcal{F}^{-1} \left\{ \mathcal{F} \left\{ \mysp{X} \right\} ^x \odot \mathcal{F} \left\{ \mysp{Y} \right\} ^y \right\}$.

In the most general form, an SSP encoding is defined as
\begin{equation}
    \phi(\myvec{x}) = \mathcal{F}^{-1} \left\{ e^{i \Theta \frac{\myvec{x}}{l}} \right\}
\end{equation}
where $\Theta \in \mathbb{R}^{D \times N}$ is the phase matrix of the SSP encoding and $l \in \mathbb{R}$ is a length-scale parameter. The phase matrix defines the mapping and ensures all SSPs generated are real-valued. Particular choices of $\Theta$ produce grid-cell-like representations \citep{dumont2020gridcell}.

SSPs can also be interpreted as probability representations over the feature space \citep{furlong2022fractional,furlong2023}. This is because similarity induces a kernel function \citep{frady2021,voelker2020}: $\phi(\myvec{x}_1) \mysim \phi(\myvec{x}_2) = k(\phi(\myvec{x}_1), \phi(\myvec{x}_2))$. The behaviour of this kernel function, $k(\cdot, \cdot)$, can be modified by changing the distribution from which the phase matrix of the encoding is sampled. Other HRR operations also behave specially on SSPs. Binding is addition in the feature space: $\phi(\myvec{x}_1) \mybind \phi(\myvec{x}_2) = \phi(\myvec{x}_1 + \myvec{x}_2)$. Note that this makes shifting items easy and means the zero vector in the feature space (i.e., the origin) is represented as the identity vector in the SSP space (i.e., $\phi(\myvec{0}) = \myvec{I}$). This is because $\phi(\myvec{x}) = \phi(\myvec{x} + \myvec{0}) = \phi(\myvec{x}) \mybind \phi(\myvec{0})$ and $\phi(\myvec{x}) = \phi(\myvec{x}) \mybind \myvec{I}$. Inversion is negation in the feature space: $\myinvert{\phi(\myvec{x})} = \phi(-\myvec{x})$. This is because $\mathbf{I} = \phi(\myvec{0}) = \phi(\myvec{x} + (-\myvec{x})) = \phi(\myvec{x}) \mybind \phi(-\myvec{x})$ and $\myvec{I} = \phi(\myvec{x}) \mybind \myinvert{\phi(\myvec{x})}$.

SSPs can be combined with other vectors to represent complex scenes. For example, we can encode $k$ items at certain positions in a scene as $\myvec{a} = \sum_{i}^{k}{\mysp{ITEM}_i \mybind \phi({\myvec{x}_i})}$. As in the discrete list example, the position of a known item in the scene or the item at a known position in the scene can be queried.

\subsection{Dual Process Cognition}

We often explain the operation of our solver with metaphors to dual process theory \citep{wason1974}, supporting cognitive plausibility. To elucidate these comparisons, we introduce dual process theory and its applications to both psychology and AI.

\subsubsection{System 1 and System 2}

In psychology, the mind is often understood as being composed of two complementary systems, System 1 and System 2 \citep{kahneman}. System 1 performs fast, automatic tasks, and can be thought of as a fictitious agent modelled as a large associative memory tuned by evolution and experience. System 1 operates constantly and involuntarily, appraising the environment to invoke emotions and produce impressions that generate a coherent world model. The operation of System 1 is characterized by heuristics, accurate but imperfect intuitions that intelligently guide behaviour. System 2 performs slow, intentional tasks, and can be thought of as another fictitious agent aligning with our conscious experience. System 2 operates sparingly and intentionally, solving complex problems only when needed because mobilizing the required cognitive resources is expensive. The operation of System 2 is characterized by calculated reasoning for attention-intensive tasks. System 1 and System 2 each play a role in intelligence because they solve different problems, so it is often argued that a complete account of intelligence must model both \citep{turing_dl}.

\subsubsection{Symbolic and Connectionist Intelligence}

Early approaches to AI \citep[e.g.,][]{dendral,eliza}, inspired by formal reasoning, sought to emulate symbolic cognition \citep{symbolic1,symbolic2}. In this paradigm, intelligence arises from strictly following explicit, hand-crafted rules. The resulting so-called expert systems excel at conveying narrow domain knowledge and performing logical reasoning, but suffer from important limitations. First, explicit symbolic rules are brittle, robust to neither changes in the environment nor noisy data in raw, perceptual form \citep{nsreview4,symbolgrounding,aihistory}. Second, discrete search is not scalable, quickly becoming intractable due to inevitable combinatorial explosion \citep{aihistory}. Third, learning is difficult, limiting generalizability and constraining systems to narrow scopes of expertise \citep{nsreview5,nsreview1}. Fourth, biological implementation of the symbolic paradigm remains unclear.

Recent successes in AI \citep[e.g.,][]{imagenet_winner,gpt2,alphago}, inspired by the observation of countless neurons activating in concert in the brain \citep{mcculloch_pitts,rosenblatt_perceptron}, have been dominated by the connectionist view \citep{pdp}. In this paradigm, intelligence arises from operations computed as functions of high-dimensional feature representations in large-scale neural networks \citep[NNs;][]{hinton_dl,schmidhuber_dl}. The resulting neural and statistical machines are extremely powerful at extracting patterns from data, but suffer from important limitations of their own \citep{fodor1988}. First, connectionist models scale to a fault, performing better with increased model and dataset size but requiring large models and datasets to succeed \citep{nsreview4,nsreview5}. Second, connectionist models learn to a fault, sometimes memorizing their training data and often generalizing poorly outside of their training distribution \citep{nsreview4,nsreview5,advex2,garymarcus,advex1}. Third, learned high-dimensional representations and operations are opaque, rendering many connectionist models unexplainable black boxes \citep{nsreview4,nsreview5,nsreview1,nsreview2}. Fourth, biological implementation of the connectionist paradigm also remains unclear \citep{connectionism_implausible,backprop_implausible}.

Connectionist models excel at System 1 tasks, such as recognizing images, transcribing speech, and generating text. Conversely, symbolic models excel at System 2 tasks, such as logical inference and trial-and-error search. We believe neurosymbolic models integrating elements of each paradigm are necessary to completely model intelligence.

VSAs provide one type of neurosymbolic approach. VSAs bridge symbolic and connectionist AI, combining the strengths of each to mitigate the weaknesses of each. Symbol-like representations afford the benefits of explicit, hand-crafted rules while overcoming the limitations of pure symbolic models. Using distributed representations provides robustness to noise, facilitates similarity metrics to guide search, and enables continuous learning. Neural-like representations afford the benefits of neural learning in NNs while overcoming the limitations of pure connectionist models. Using structured representations with useful inductive biases and features pre-encoded improves sample-efficiency in learning, enables generalizable algebraic models, and permits direct interpretation of intermediate representations. Additionally, VSAs are biologically and cognitively plausible and can be implemented in spiking NNs \citep{eliasmith2003neural,eliasmith2013how}.

\subsection{Abstraction and Reasoning Corpus}

Before explaining how our solver works, we provide a more detailed description of ARC-AGI. Here, we discuss its theoretical foundations, its exact format, how others have approached it, and how we approach it.

\subsubsection{Background and Definition}

AI was originally defined as ``the science of making machines do things that would require intelligence if done by [people]'' \citep{minsky}. This skill-based view of intelligence underpins many modern AI systems adept at one particular task, but only that one task. The skills AI systems can perform---synthetic art generation and natural language translation, for instance---have become increasingly advanced, but the underlying paradigm remains the same: models fit their many parameters to copious amounts of data during a training phase, and statically apply their learned skill to unseen data during an inference phase. In contrast, humans are simultaneously adept at many tasks. A skilled human may be able to both create art and translate between languages, and unskilled humans can learn these new skills. In this sense, human intelligence is better characterized by its adaptability, fluidity, and generality. Not only can we humans excel at particular well-known tasks, we can rapidly learn to perform new tasks in response to new situations. Human intelligence is dynamic, without disjoint training and inference phases. In ARC-AGI's initial introduction, Chollet defined intelligence as skill-acquisition efficiency: the ability of a system to quickly develop novel skills to solve novel problems \citep{chollet2019measureintelligence}. In this view, crystalline skill at any number of tasks is not intelligence; instead, intelligence is the process by which those skills are learned. ARC-AGI aligns with this paradigm by testing the ability of a system to learn arbitrary transforms on-the-fly from few examples.

ARC-AGI is a collection of abstract reasoning \textit{tasks} (see Figs. \ref{fig:task_0962bcdd} and \ref{fig:task_54d9e175}). The original dataset, ARC-AGI-1, consists of 400 public training tasks, 400 public evaluation tasks, and 200 private evaluation tasks. The newer version, ARC-AGI-2, consists of 1000 public training tasks, 120 public evaluation tasks, and 120 private evaluation tasks. The performance of a solver is usually assessed on each group separately because the evaluation splits are more difficult. The private splits are known only to the ARC-AGI team and are used to test the true generalizability of solvers. Each task has a \textit{training} component and a \textit{testing} component. The training component is a set, $\mathcal{D}$, of a few \textit{demonstrations}, or demonstration pairs. The number of demonstrations is a feature of the task, but is usually about three (in ARC-AGI-1-Train, $2 \leq |\mathcal{D}| \leq 10$ with median $3$, mean $3.26$, and standard deviation $0.96$). Each demonstration is composed of two \textit{grids}, one \textit{input} grid and one \textit{output} grid. Each grid, $\mathbf{G} \in \gridspace$, contains $r$ rows and $c$ columns of \textit{pixels}. Grids must be rectangular; they are often square, but $r$ need not equal $c$. Grids range in size (in ARC-AGI-1-Train, $r \text{, } c \in \{1, \ldots, 30\}$ with medians $10$ and $10$, means $9.64$ and $10.03$, and standard deviations $6.01$ and $6.16$); the sizes of the input and output grids in a demonstration often match, but a grid size change may be part of the task. Each pixel, $P_{ij} \in \pixelspace$, is one of ten discrete symbols provided as an integer but usually displayed as a colour for human convenience. The exact colours used for display are arbitrary; no task relies on facts about colours, such as that orange is a mix of red and yellow. To summarize, $\mathcal{D} = \{ (\mathbf{G}_{I_d}, \mathbf{G}_{O_d}) ~ | ~ \mathbf{G}_{*_*} \in \gridspace, ~ d = 1 \ldots |\mathcal{D}| \}$, where $\gridspace = \pixelspace^{r_{*_*} \times c_{*_*}}$ and $\pixelspace = \{ 0, \ldots, 9 \}$. The input grid of each demonstration can be mapped onto its corresponding output grid via some \textit{transformation}. This transformation, $\mathcal{T}_d$, describes how to produce this specific output grid, $\mathbf{G}_{O_d}$, from the corresponding input grid, $\mathbf{G}_{I_d}$. The transformations underlying each demonstration in a task share structure and can be generalized into a simple common \textit{program}. This program, $\mathcal{P}$, describes how to produce the corresponding output grid for any valid input grid. To summarize, $\mathbf{G}_{I_d} \xrightarrow{\mathcal{T}_{d}} \mathbf{G}_{O_d}$ for each $d \in \{1, \ldots, |\mathcal{D}|\}$ and $\mathbf{G}_{I_d} \xrightarrow{\mathcal{P}} \mathbf{G}_{O_d} ~~ \forall d \in \{1, \ldots, |\mathcal{D}|\}$. The testing component is also a set, $\mathcal{Q}$, of a few \textit{queries}. The number of queries is also a feature of the task, but is usually just one (in ARC-AGI-1-Train, $1 \leq |\mathcal{Q}| \leq 3$ with median $1$, mean $1.04$, and standard deviation $0.22$). A query is a lone test input grid without a provided output grid. The goal of the solver is to correctly answer the queries; to solve a task correctly, the solver must predict exactly the correct output grid, both its size and contents, for all queries. The performance of a solver is defined as the proportion of some subset of tasks solved correctly.

What makes ARC-AGI so difficult is also what makes it compelling. There are several important differences between ARC-AGI and other benchmarks. First, the problem is generative. Instead of choosing the correct output grid from a set of candidates as in classification tasks, such as the RPMs, solvers must generate the output grid from scratch. Second, the reasoning involved is, as the name suggests, abstract. Implied objects are not grounded (i.e., they may have no real-world correlates), implied actions invoke high-level ideas (e.g., motion until contact), and implied rules invoke relative and fuzzy concepts (e.g., the largest among a set or the horizontalness of a shape). As such, solvers must represent each grid in terms of its abstract qualities and construct a program insensitive to inconsequential details of the particular demonstrations. Third, no two tasks are alike. No solution program will work for any pair of tasks, and the possible concepts, transforms, and their compositions are wildly unconstrained.

Despite these complexities, ARC-AGI is based only on four core knowledge priors: objectness, goal-directedness, numbers, and geometry. Objectness requires perceiving grids as collections of cohesive, persistent, interacting objects. Goal-directedness requires anthropomorphizing objects into agents acting to achieve goals. Numbers requires understanding basic arithmetic, such as counting, addition, and subtraction, and abstract mathematical notions, such as sorting, smallest, and largest. Geometry requires understanding position, symmetry, translation, rotation, scaling, and the like. A solver with a thorough understanding of only these four concepts can solve ARC-AGI; specialized knowledge, such as language or external world facts, are unnecessary. Humans innately have some of these priors from evolution, and acquire the others through experience interacting with the modern world.

\subsubsection{Related Work}

Many ARC-AGI solvers can be divided into two broad classes: \textit{transductive} solvers and \textit{inductive} solvers \citep{transind,arcprize2024}. Transductive solvers \citep[e.g.,][]{testimetuning,nnarc,jackcole,miniarc_solver,architects,2dngpt,hrm_arc} directly answer the queries without explicit intermediate reasoning. These approaches use an NN, usually an LLM, to implicitly understand the demonstration transformations and apply the query transformations all at once, without generating any representation of the patterns underlying the transformations. Inductive solvers \citep[e.g.,][]{objmdl,relationaldecomposition,abductivesymbolsolver,nonprocedural,ilp,icecuber,dividealignconquer,arga} generate explicit rules mapping input grids to output grids and apply these rules to the queries. These approaches use various search techniques, but invariably cast ARC-AGI as a program synthesis problem. Both approaches in isolation are problematic; neurosymbolic approaches \citep[e.g.,][]{dreamcoder_arc,nssolver,bonnetmacfarlane,ouellette,vaearithmetic,llms_hypotheses_arc} integrating transduction and induction, such as deep-learning-guided program synthesis, have been suggested as the ideal solution \citep{arcprize2024}.

Transductive solvers are dominated by LLMs and other transformer-based \citep{transformer} models. But, despite their impressive abilities, LLMs suffer from several critical weaknesses that prove problematic for solving ARC-AGI. First, LLMs struggle with numeracy. Stemming from fundamental limitations of the transformer architecture \citep{llms_counting_1,llms_counting_2}, LLMs struggle to perform out-of-distribution arithmetic \citep{llms_arithmetic} or, infamously, to even count the number of occurrences of a letter in a sequence \citep{llms_letters}. Second, LLMs struggle with following rules. As such, LLMs cannot play many games effectively \citep{llms_childplay}, and even their `reasoning' variants cannot follow steps well enough to consistently solve complex problems \citep{lrms_rules}. Third, LLMs struggle with out-of-distribution generalization. LLMs lack robustness and fail to exhibit even simple forms of generalization: they cannot generalize facts into their reversed form \citep{llms_reversalcurse}, are biased by content effects \citep{llm_contenteffects}, and brittly collapse on slight variations of known tasks \citep{llms_counterfactual,llms_caesar}. Fourth, LLMs struggle with important capabilities of human intelligence. LLMs cannot learn fast, generalize broadly, or explicitly understand their actions \citep{sudoku}. Language is useful to express reasoning, but is not required to perform reasoning \citep{lanaguge_thought_nature}. These failures of LLMs, along with their restriction to the text domain, suggest deficiencies in ARC-AGI's required core knowledge priors and intrinsic issues with performing the systematic reasoning required to solve ARC-AGI. LLMs, when prompted directly or with chain-of-thought, and vision language models in isolation have weak performance \citep{arc_llmiq,loth,gpt4v_arc,gpt4vstructuredreasoning}.

Inductive solvers perform search on either a general-purpose language (GPL) or a domain-specific language (DSL). GPLs, such as Python, are expressive and can be used to solve all ARC-AGI tasks, providing scalability. But, GPLs lack the inductive biases necessary to express solutions to ARC-AGI simply, making search difficult. Conversely, DSLs, requiring hand-crafted primitives, allow humans to introduce problem-relevant inductive biases, making search easier. But, DSLs are not expressive enough to solve all possible ARC-AGI-like tasks, limiting scalability. DSLs capable of solving all existing ARC-AGI tasks \citep{hodel_arcdsl,hodel_rearc} are massive and still may be incapable of solving new tasks. Additionally, without proper guidance through the search space, search processes are unintelligent.

Unfortunately, little is known about how exactly humans solve ARC-AGI so effectively. But, some facts are clear from studies on humans. First, humans reason in terms of objects. This is true in general, and action traces of humans solving ARC-AGI tasks indeed show pixel groups are operated on together \citep{fastflexible}. Second, humans dedicate time to explicit hypothesis generation. A non-negligible period of initial inaction is consistently observed \citep{fastflexible}. Third, humans can express their task solutions programmatically. Transformations can be communicated in terms of natural language rules \citep{tenenbaumlarc}; whether the solution discovery process itself is program synthesis remains unknown. Fourth, humans conceptualize features and operations using high-level abstractions. For example, humans describe their solutions in terms of objects with ``tails'' \citep{fastflexible}, objects as ``flowers'' \citep{harcupdate}, and objects ``bumping into'' each other \citep{tenenbaumlarc}; whether such abstractions are necessary for the solving process remains unknown. Fifth, humans make mistakes different from those made by artificial solvers. Human failures usually show partial correctness indicating some understanding \citep{harc,harcupdate}, but the failures of artificial solvers indicate little understanding \citep{kidsarc}. We take these facts to suggest humans solve ARC-AGI by means of object-centric program synthesis.

Importantly, transductive and inductive solvers are each doing something entirely unlike what humans are doing; humans are not solving ARC-AGI purely by means of transduction or induction. Transductive solvers, based on connectionist networks, model System 1; inductive solvers, based on symbolic engines, model System 2. But, neither is enough alone: humans are able to solve ARC-AGI so effectively because of contributions from both System 1 and System 2. Humans are guided by System 1 intuitions, immediate instincts arising from common sense and world knowledge, but also generate and test explicit hypotheses with System 2 reasoning, careful consideration of a limited number of ideas. We believe a neurosymbolic approach combining the strengths of symbolic and connectionist AI is thus needed.

\section{Methods}
\label{sec:methods}

Fundamentally, we aim to develop a cognitively plausible ARC-AGI solver. The goal of ARC-AGI is to spur progress towards artificial general intelligence (AGI), so taking inspiration from human intelligence, the only known general intelligence, is reasonable. Additionally, human intelligence remains the best ARC-AGI solver by a wide margin, suggesting important insights are missing from artificial solvers.

\subsection{Solution Structure}

We believe humans solve ARC-AGI via object-centric program synthesis. As such, our solver generates solutions to ARC-AGI tasks based on objects and programs.

\subsubsection{Objects}

Humans reason about ARC-AGI with \textit{objects} rather than with individual pixels or whole grids. Fundamentally, an object is a group of pixels transformed cohesively. Because the transformations are different in every task, what makes a group of pixels constitute an object is task-dependent; the same group of pixels may form an object in one task, but not in another. Thus, for each task, an object definition must be discovered alongside the transformations as part of the solution generation process. Following this view, our solver represents each grid, input and output, as the set of its constituent objects and each transformation, training and testing, as a series of one-to-one operations on those objects. 

\begin{figure}[t]
    \centering
    \begin{minipage}{0.47\textwidth}
        \centering
        \includegraphics[width=\linewidth]{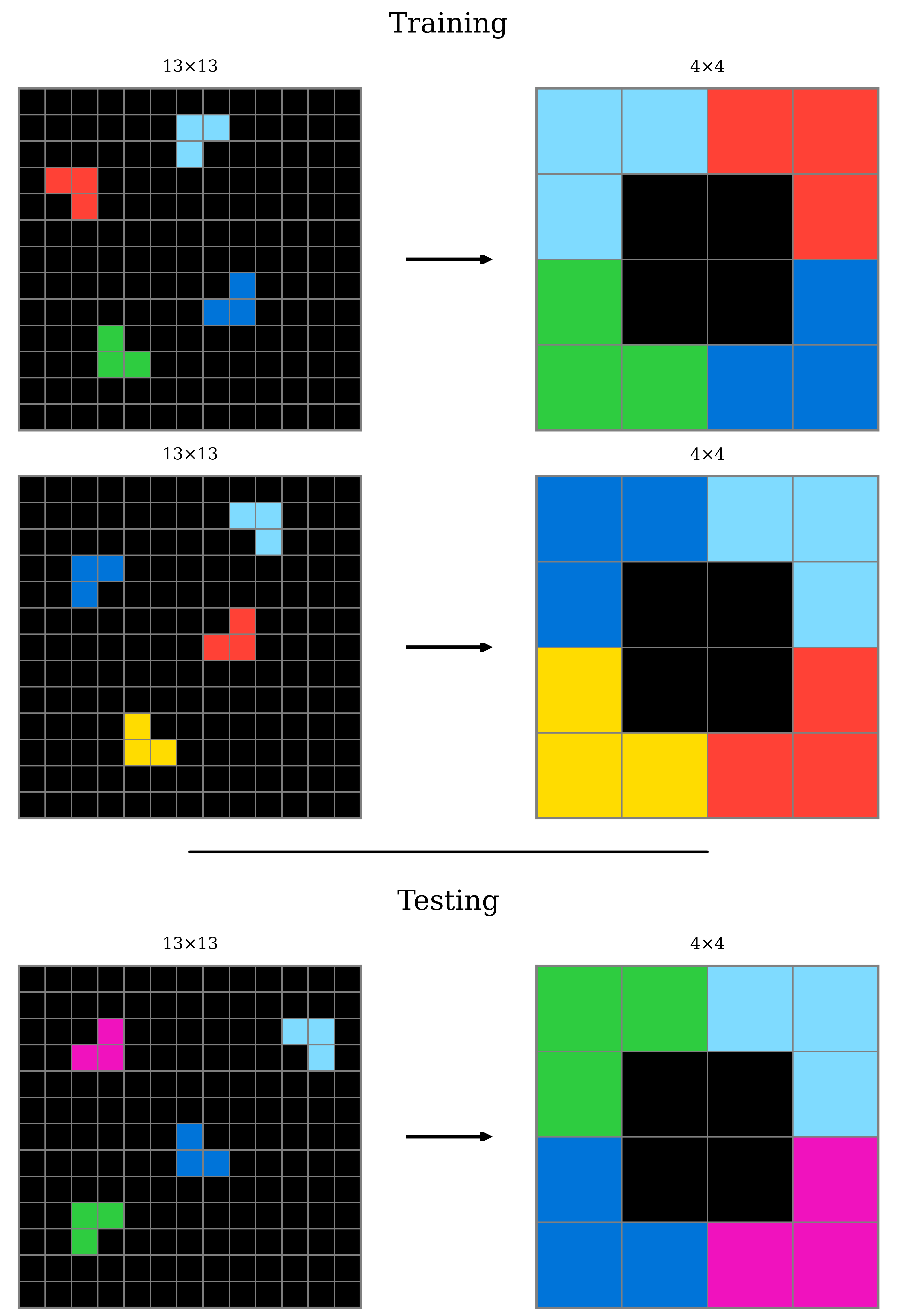}
        \caption{ARC-AGI task \texttt{a61ba2ce}. Here, the implicit pattern is to rearrange the four blocks into a hollow square on a small grid.}
        \label{fig:task_a61ba2ce}
    \end{minipage}
    \hfill
    \begin{minipage}{0.47\textwidth}
        \centering
        \includegraphics[width=\linewidth]{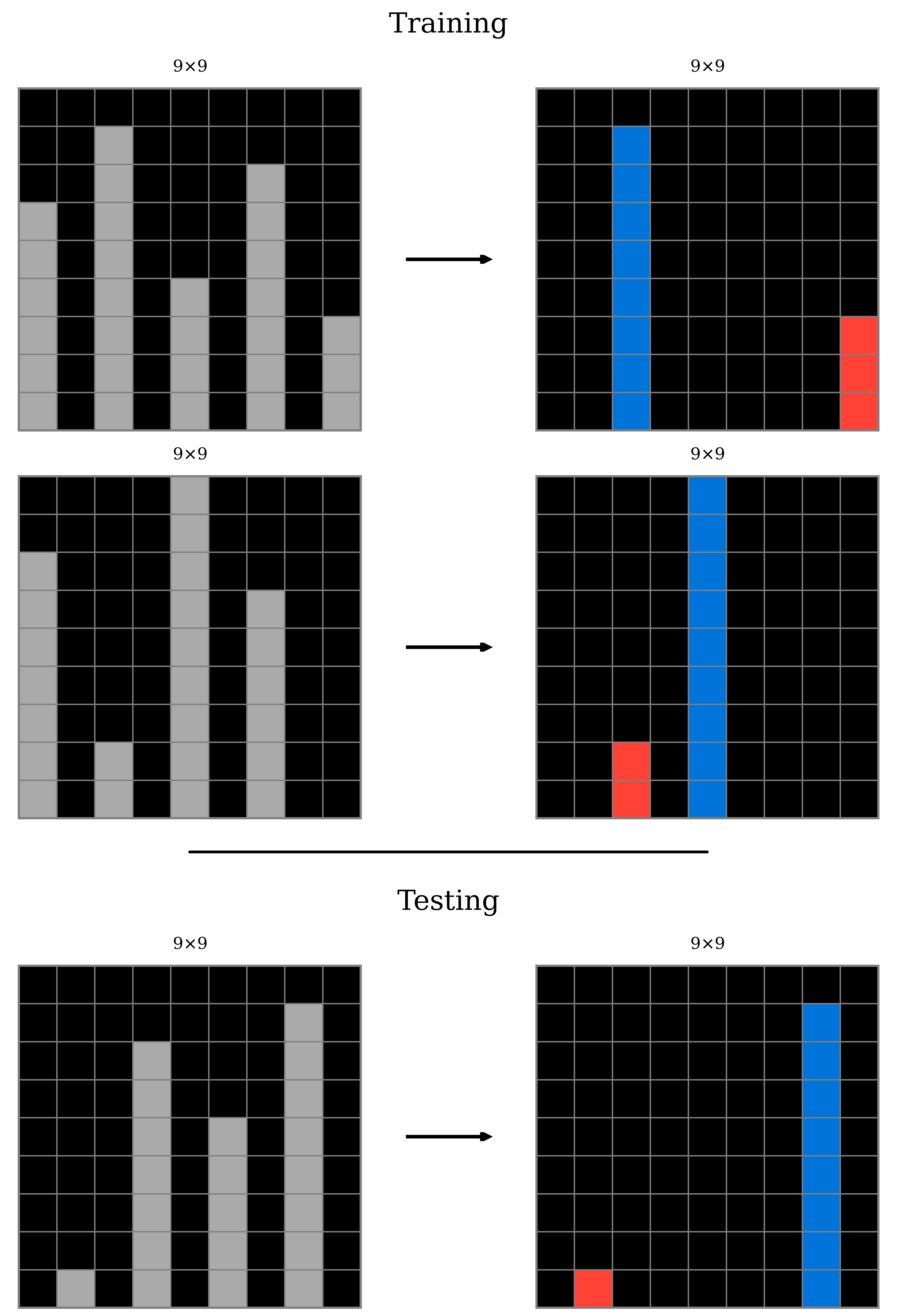}
        \caption{ARC-AGI task \texttt{a61f2674}. Here, the implicit pattern is to recolour the shortest stripe red and the tallest stripe blue.}
        \label{fig:task_a61f2674}
    \end{minipage}
\end{figure}

Object representation determines both what tasks can be solved and how those tasks can be solved. Certain object properties may be required to solve some tasks, but not others. For example, an object's colour is relevant in ARC-AGI tasks \texttt{54d9e175} and \texttt{a61f2674} (see Figs. \ref{fig:task_54d9e175} and \ref{fig:task_a61f2674}), but irrelevant in ARC-AGI tasks \texttt{0962bcdd} and \texttt{a61ba2ce} (see Figs. \ref{fig:task_0962bcdd} and \ref{fig:task_a61ba2ce}). Additionally, the way information is represented makes some transformations easier to implement, but others harder. For example, representing position in Cartesian coordinates makes performing translations simple but rotations complicated, and vice versa for polar coordinates.

\begin{figure}
    \centering
    \includegraphics[width=0.6\linewidth]{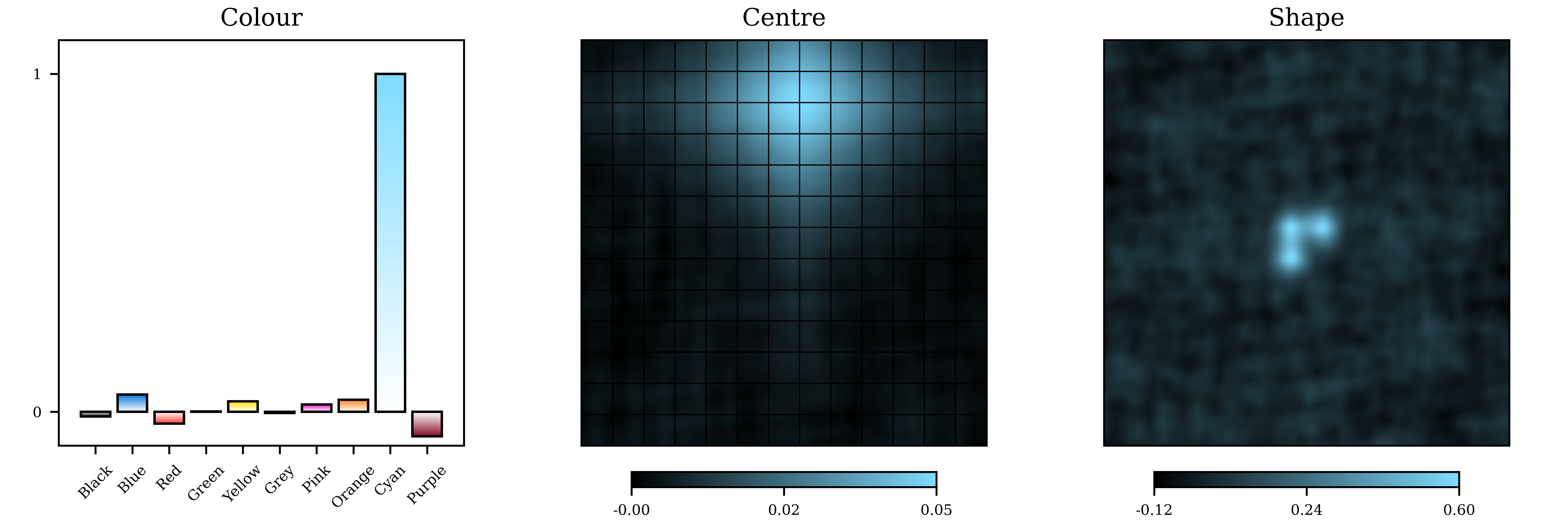}
    \includegraphics[width=0.6\linewidth]{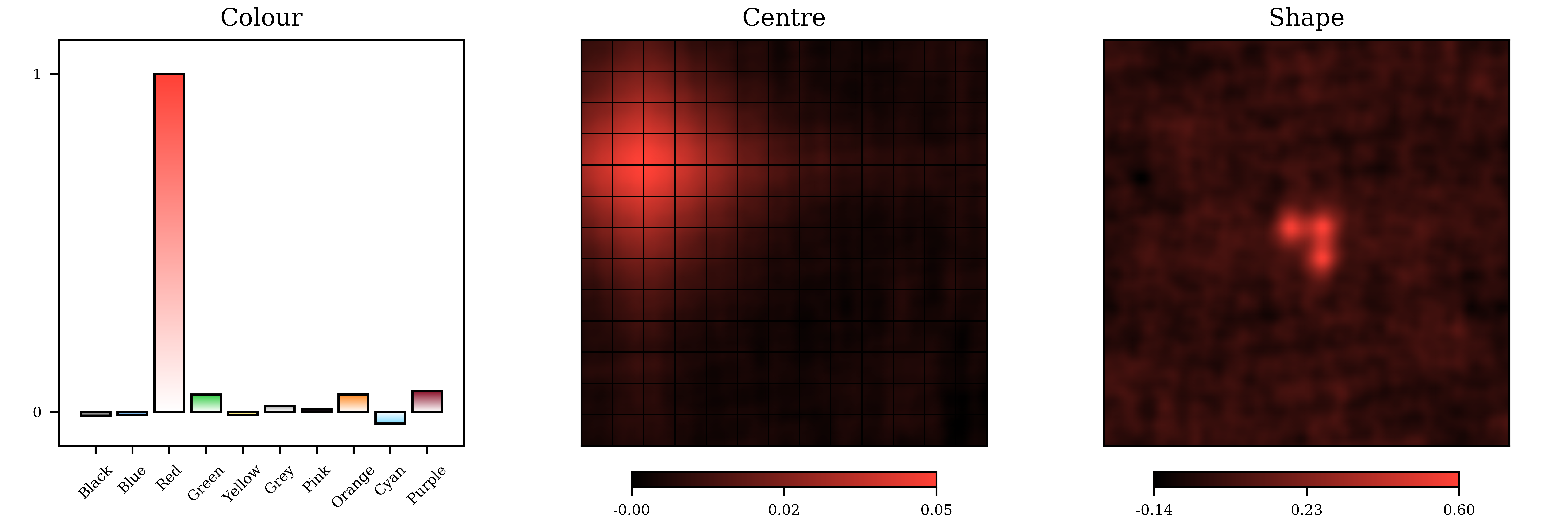}
    \includegraphics[width=0.6\linewidth]{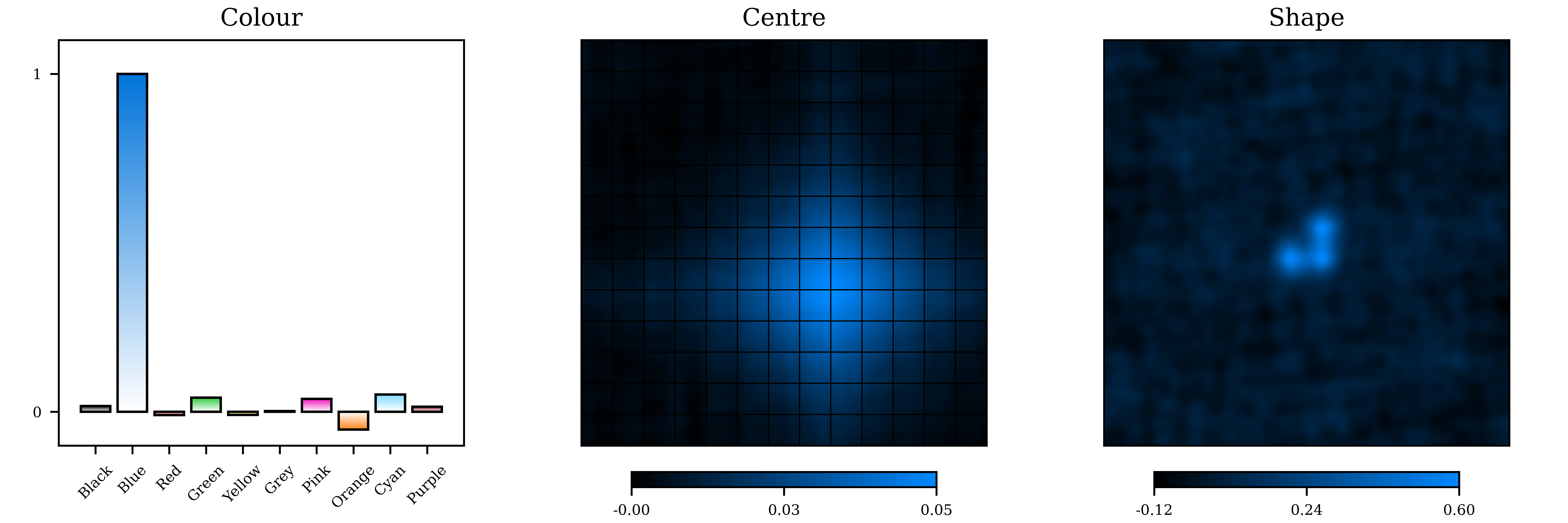}
    \includegraphics[width=0.6\linewidth]{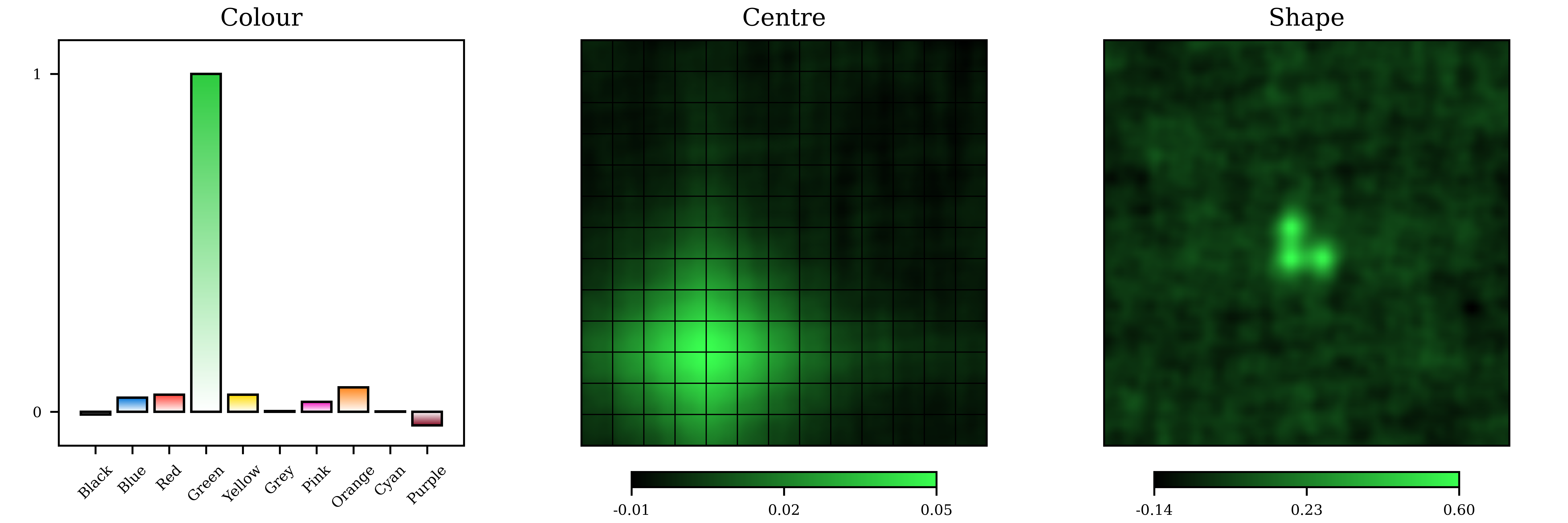}
    \caption{Visualization of the object representations for the input grid in the first demonstration of ARC-AGI task \texttt{a61ba2ce} (see Fig. \protect{\ref{fig:task_a61ba2ce}}). The colour subfigure (left) shows the similarity of the object's colour representation to each of the ten possible colour vectors. The centre subfigure (middle) shows the similarity of the object's centre representation to the SSP encoding each location in the grid. The shape subfigure (right) shows the similarity of the object's shape representation to the SSP encoding each possible location in two-dimensional space. These representations use $N \! = \! 1024$-dimensional vectors.}
    \label{fig:a61ba2ce_objects}
\end{figure}

We adopt an object representation scheme with three features: \textit{colour}, \textit{centre}, and \textit{shape}. An object's colour is its discrete symbol in the grid. Colour is represented as one of ten vectors; as such, objects may only be one colour. An object's centre is the midpoint of the furthest extent of all its pixels in all directions. Centre is represented as the SSP encoding the two-dimensional coordinates of this midpoint relative to the grid's centre, with a blurring effect facilitating partial similarity. An object's shape is its pixel pattern. Shape is represented as the normalized bundle of the SSPs encoding each pixel's location relative to the object's centre. Object representations can be visualized using standard SPA and SSP methods (see Fig. \ref{fig:a61ba2ce_objects}). We hypothesize that higher-level object properties, such as symmetry and exact pixel count, initially go unnoticed by humans and are only encoded after their relevance becomes apparent. Although necessary to rigourously solve some tasks, these properties are not explicitly considered by our solver.

\subsubsection{Programs}

Humans solve ARC-AGI with \textit{programs}. In this view, solving an ARC-AGI task means synthesizing a general program to generate the correct output grid for any valid input grid. Crucially, the search process underlying program discovery is both constrained and intentional. To solve an ARC-AGI task, humans iteratively explore multiple solution hypotheses---re-interpreting the contents of the grids and conceptualizing new transformations based on different abstractions---until they find a satisfactory algorithm to solve all demonstrations. However, the extent of this exploration is narrow; because System 2 cognitive resources are limited, humans consciously examine only a few solution hypotheses. System 1 intuitions guide which paths are explored; humans quickly dismiss nonsensical ideas and only seriously investigate ideas that conceptually fit the task's content. Following this view, our solver iteratively constructs and tests hypotheses represented as solution programs.

Program representation also determines both what tasks can be solved and how those tasks can be solved. Our solver synthesizes programs within a DSL. The primitives in this DSL were carefully designed to capture human inductive biases commonly appearing in ARC-AGI while remaining as general-purpose as possible. For example, our DSL contains a basic \myoperation{Identity} operation to exactly reproduce an input object as an output object, elementary \myoperation{Recolour}, \myoperation{Recentre}, and \myoperation{Reshape} operations to change each object property, an open-ended \myoperation{Generate} operation to create an arbitrary output object, and more (see Table \ref{tab:dsl_primitives}).

\begin{table}[t]
    \caption{Our solver's DSL.}
    \begin{center}
        \begin{tabular}{p{0.15\linewidth}p{0.15\linewidth}p{0.5\linewidth}}
            \toprule
            \multicolumn{1}{c}{Operation} & \multicolumn{1}{c}{Parameter(s)} & \multicolumn{1}{c}{Explanation} \\
            \midrule
            \myoperation{Identity} & --- & Keeps the object as is. \\
            \myoperation{Extract} & --- & Shrinks the grid around the object. \\
            \myoperation{Recolour} & $\mysp{COLOUR}$ & Changes the object's colour to $\mysp{COLOUR}$. \\
            \myoperation{Recentre} & $\mysp{CENTRE}$ & Changes the object's centre to $\mysp{CENTRE}$. \\
            \myoperation{Reshape} & $\mysp{SHAPE}$ & Changes the object's shape to $\mysp{SHAPE}$. \\
            \myoperation{Move} & $\mysp{AMOUNT}$ & Shifts the object's centre by $\mysp{AMOUNT}$. \\
            \myoperation{Gravity} & $\mysp{DIRECTION}$ & Shifts the object's centre in $\mysp{DIRECTION}$ until contact with another object or grid boundaries. \\
            \myoperation{Grow} & $\mysp{DIRECTION}$ & Changes the object's centre and shape by stretching the object in $\mysp{DIRECTION}$ until contact with another object or grid boundaries.  \\
            \myoperation{Fill} & --- & Fills in the object's shape. \\
            \myoperation{Hollow} & --- & Hollows out the object's shape. \\
            \myoperation{Generate} & $\mysp{COLOUR}$, $\mysp{CENTRE}$, $\mysp{SHAPE}$ & Creates an object with colour $\mysp{COLOUR}$ and shape $\mysp{SHAPE}$ at centre $\mysp{CENTRE}$. \\
            \bottomrule
        \end{tabular}
    \end{center}
    \label{tab:dsl_primitives}
\end{table}

\begin{figure}
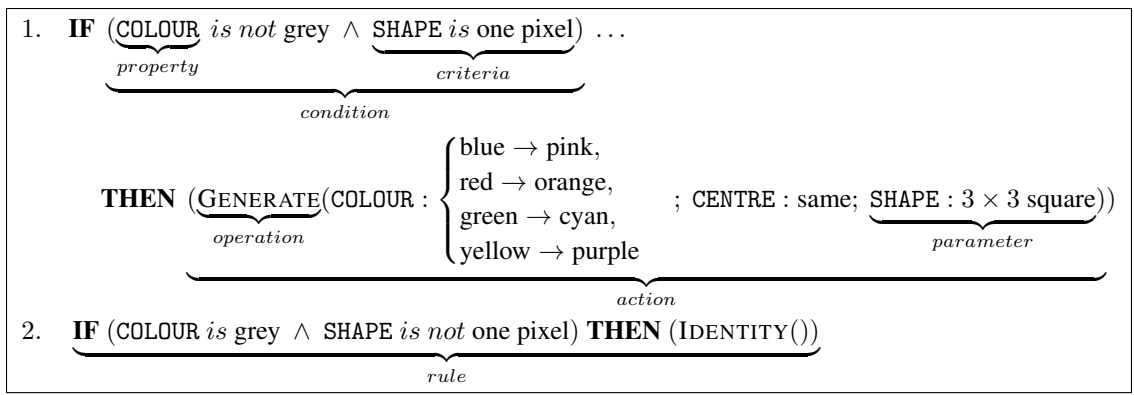

    \centering
    \begin{equation*}
        \boxed{
            \begin{aligned}
                & ~ 1. ~~~~ \textbf{IF} ~ \underbrace{( \underbrace{\mysp{COLOUR}}_{property} ~ is ~ not ~ \text{grey} ~ \land ~ \underbrace{\mysp{SHAPE} ~ is ~ \text{one pixel}}_{criteria} )}_{condition} ~ \ldots \\
                & ~~~~~~~~~~~~~ \textbf{THEN} ~ \underbrace{( \underbrace{\myoperation{Generate}}_{operation} ( \mysp{COLOUR} : \begin{cases}
                    \text{blue} \rightarrow \text{pink}, \\
                    \text{red} \rightarrow \text{orange}, \\
                    \text{green} \rightarrow \text{cyan}, \\
                    \text{yellow} \rightarrow \text{purple} \\
                \end{cases} ; ~ \mysp{CENTRE} : \text{same} ; ~ \underbrace{\mysp{SHAPE} : 3 \times 3 ~ \text{square}}_{parameter} )}_{action} ) ~ \\
                & ~ 2. ~~~~ \underbrace{\textbf{IF} ~ ( \mysp{COLOUR} ~ is ~ \text{grey} ~ \land ~ \mysp{SHAPE} ~ is ~ not ~ \text{one pixel} ) ~ \textbf{THEN} ~ ( \myoperation{Identity} () )}_{rule} \\
            \end{aligned}
        }
    \end{equation*}
    \caption{One possible solution program for ARC-AGI task \texttt{54d9e175} (see Fig. \protect{\ref{fig:task_54d9e175}}).}
    \label{fig:54d9e175_solution}
\end{figure}

More formally, our solver produces a solution to each task as a \textit{program} (see Fig. \ref{fig:54d9e175_solution} for an example). This program can be instantiated to generate the correct transformation for each demonstration and, ideally, unseen valid input grids. We define a solution program as a set of \textit{rules}. Each rule describes a different type of object-to-object mapping as an if-then statement expressing an \textit{action} to be taken based on some \textit{condition}. Conditions are logical compositions of \textit{criteria}. Criteria reflect some state or feature of a particular object \textit{property}. For example, the condition for a rule may be ``$\mysp{COLOUR} ~ is ~ not ~ \text{grey} ~ \land ~ \mysp{SHAPE} ~ is ~ \text{one pixel}$'', a conjunction of the criteria ``$\mysp{COLOUR} ~ is ~ not ~ \text{grey}$'' and ``$\mysp{SHAPE} ~ is ~ \text{one pixel}$'' based on the $\mysp{COLOUR}$ and $\mysp{SHAPE}$ properties of an object. Conditions may be vacuous, causing the associated action to apply to all objects. Actions are particular \textit{operations}, taken from our DSL, to apply to each object satisfying the rule's condition. Operations convert one input object into one output object in some structured way. Open-ended operations require \textit{parameters}. Parameters determine how each instantiation of the operation should behave. For example, the action for a particular rule may be ``\myoperation{Generate}$( \mysp{COLOUR} : \text{orange}; ~ \mysp{CENTRE} : \text{origin}; ~ \mysp{SHAPE} : 3 \times 3 ~ \text{square} )$'', representing the application of the \myoperation{Generate} operation with the $\mysp{COLOUR}$ parameter as orange, $\mysp{CENTRE}$ parameter as the origin, and $\mysp{SHAPE}$ parameter as a $3 \times 3$ square.

\subsection{Solution Synthesis}

Our solver models human cognition throughout all stages of the solving process. For example, humans often start solving ARC-AGI tasks by sizing the output grids to be generated \citep{fastflexible,harc}. As such, our solver begins by generating a hypothesis about how the grid size changes in the task (see Table \ref{tab:size_hypotheses}). If the sizes of the input and output grid are the same in each demonstration, then the size of each test output grid is predicted to the be the same as the corresponding query; otherwise, if the sizes of all output grids across all demonstrations are the same, then the size of each test output grid is predicted to be the same as this constant size; otherwise, the output grid sizes are determined by the actions in the solution program.

\begin{table}[b]
    \caption{Our solver's grid size change hypotheses.}
    \begin{center}
        \begin{tabular}{p{0.15\linewidth}p{0.6\linewidth}}
            \toprule
            \multicolumn{1}{c}{Hypothesis} & \multicolumn{1}{c}{Description} \\
            \midrule
            \myoperation{Identity} & Each output grid is the same size as its corresponding input grid. \\
            \myoperation{Constant} & All output grids are of the same constant size. \\
            \myoperation{Function} & The size of the output grid is some function of the size of the input grid or its contents. \\
            \bottomrule
        \end{tabular}
    \end{center}
    \label{tab:size_hypotheses}
\end{table}

We believe humans solve ARC-AGI using multiple types of logical reasoning. As such, we divide the rest of our solver's solution generation process into three stages, each completed by one mode of reasoning: \textit{demonstration abduction}, \textit{rule induction}, and \textit{answer deduction}. First, demonstration abduction generates an object definition and a set of simple steps to explain each demonstration (i.e., the objects and actions to produce each training output grid from its corresponding input grid). Simply understanding what is happening in each demonstration is necessary for further reasoning; abductive reasoning can find the most probable, usually the simplest, causal explanation for these observations. Second, rule induction generates broadly-applicable rules expressing when and how actions are applied, accounting for all observations. Finding the commonalities and distinctions underlying all observed action applications is necessary for true understanding; inductive reasoning can generalize these specific observations into rules. Third, answer deduction applies the rules generated to each query to produce a corresponding output grid. Whatever understanding of the demonstrations has been developed must be applied to the novel test input grids; deductive reasoning can draw logically sound inferences from these well-defined premises. Together, these three steps bridge multiple reasoning modes to model the entire ARC-AGI solving process.

\subsubsection{Demonstration Abduction}

First, our solver abduces the objects and actions in the demonstrations. Any solver must conceptualize what it observes in the demonstrations before it can generalize these observations into rules. This requires determining, in parallel, what exactly constitutes an object in this particular task and what specific actions on the input objects can produce the correct set of output objects. Both the object hypothesis and action hypotheses should be as simple as possible.

Determining both an object hypothesis and action hypotheses creates a causality paradox: an object is defined as a group of pixels transformed cohesively, but actions are ill-defined without objects to transform. We solve this paradox by searching for action hypotheses based on evolving object proposals. Our solver applies an initial object hypothesis and abduces actions for these objects; only when this object hypothesis fails or yields overly complex actions are other object hypotheses considered.

We believe humans do the same. When first seeing a new ARC-AGI grid, System 1 produces initial ideas of what the objects are. This is part of the objectness core knowledge prior; humans have strong inductive biases, innate and learned, about what objects look like. These initial object proposals are often correct, but, when they fail, humans mobilize System 2 to carefully consider other object hypotheses. This top-down approach works because the space of possible object definitions is smaller than the space of possible actions. A bottom-up approach constructing objects out of individual pixels transformed cohesively, in addition to being cognitively implausible, fails because cohesive object-level actions are often not cohesive pixel-level actions. For example, a rotation transformation is simple at the object level but ill-defined at the pixel level.

Developing a sufficient set of action hypotheses requires explaining the presence of each output object in each demonstration by the application of some action to one of the corresponding input objects. Difficulty arises because usually each output object can be explained by multiple actions applied to multiple input objects. This is because our DSL has operations with inherently overlapping functionality as well as a \myoperation{Generate} operation that can represent arbitrary transformations from any input object.

There are multiple valid ways to solve each ARC-AGI task. Consider ARC-AGI task \texttt{a61ba2ce} (see Fig. \ref{fig:task_a61ba2ce}); solving this task can be conceptualized as moving each object into its corresponding corner on a smaller grid, colouring each corner of a smaller grid according to the colour of the corresponding object, and so on. For this task, our solver's approach resembles the first description. In this view, the actions needed to complete each transformation, once the output grid is resized to $4 \times 4$, are to move the upper-left-corner-shaped object to the upper left corner of the grid, move the upper-right-corner-shaped object to the upper right corner of the grid, and so on. However, even this conceptualization can be implemented as a program in multiple ways. Our DSL has both a \myoperation{Recentre} operation for absolute motion and a \myoperation{Move} operation for relative motion; each operation proves useful for certain tasks, but this task calls for absolute motion.

Consider the action that transforms the red object in the first demonstration; we can represent this as a $\myoperation{Recentre}( \mysp{CENTRE} \! : \! (1, 1) )$ or a $\myoperation{Move}( \mysp{AMOUNT} \! : \!  (5.5, -1.5) )$. Both are correct and equally valid, but abduction demands we choose the simpler explanation of this mapping. Here, we propose simplicity is determined by the relative frequency of the actions; in other words, recurring operations and parameters have greater explanatory power. The $\myoperation{Move}( \mysp{AMOUNT} \! : \!  (5.5, -1.5) )$ action explains only one of eight total output objects, but the $\myoperation{Recentre}( \mysp{CENTRE} \! : \!  (1, 1) )$ action explains two---the upper right corner object in both demonstrations. In this sense, an action set containing $\myoperation{Recentre}( \mysp{CENTRE} \! : \!  (-1, 1) )$, $\myoperation{Recentre}( \mysp{CENTRE} \! : \!  (1, 1) )$, $\myoperation{Recentre}( \mysp{CENTRE} \! : \!  (-1, -1) )$, and $\myoperation{Recentre}( \mysp{CENTRE} \! : \!  (1, -1) )$ offers a simpler explanation of the observations than any action set based on \myoperation{Move} could. The \myoperation{Recentre} operation with possible parameters $(-1, 1)$, $(1, 1)$, $(-1, -1)$, and $(1, -1)$ accounts for all observations most cohesively.

Constructing the simplest action set explaining all demonstrations can be cast as solving the minimum hitting set problem: given a set $\mathcal{S}$ of $K$ partial action sets $\mathcal{A}_k$ taken from all possible actions $\mathcal{U}$, $\mathcal{S} = \{ \mathcal{A}_k ~ | ~ \mathcal{A}_k \subseteq \mathcal{U}, ~ k = 1 \ldots K \}$, find the action set $\mathcal{A}$ hitting each partial action set in $\mathcal{S}$, $\mathcal{A} \subseteq \mathcal{U}$ and $\mathcal{A} \cap \mathcal{A}_k \neq \emptyset ~~ \forall k = 1 \ldots K$, with minimal cost according to some function $f$. Here, the cost function penalizes each additional operation and parameter introduced; in other words, simple action sets use the least number of operations and distinct operation-parameter compositions. The hitting set problem is NP-Complete, but can be exactly solved quickly in practice.

The demonstration abduction process involves iteratively attempting object hypotheses and finding the simplest action set to explain all demonstrations. This requires intelligently traversing the object hypothesis space, inferring which input object explains each output object, and inferring what actions can perform each input-output object transform. In humans, these search processes are guided by System 1 heuristics. Here, leveraging VSAs, we model such heuristics as rapid computations on the object representations. These heuristics are fast and simple, with strong but imperfect accuracy. We propose VSA-based heuristics for each of these three tasks.

The first heuristic proposes an intelligent order to traverse the object hypothesis space. Instead of randomly trying object hypotheses, our solver first tries the most promising hypotheses. Fundamentally, a useful object hypothesis produces a simple grid representation while enabling simple input-output object transforms. Simple grid representations have as few objects as possible, and simple input-output object transforms are easiest when each output object is uniquely similar to one input object. This heuristic initially encodes objects according to all six hypotheses (see Table \ref{tab:object_hypotheses}) and, for each output object, examines the distribution of the similarities to its possible corresponding input objects. We take the softmax of each similarity vector, padded with $0$ and $1$ values to ensure there are multiple elements, and consider the resultant vector's maximum value. The higher this value, the fewer competing input objects; each additional input object will have some finite similarity to the output object, reducing the value of the maximum softmax score and penalizing object hypotheses that create too many objects. Additionally, the higher this value, the more similar one input object is compared to all others; input objects with close similarities have close softmax values, preventing one singular high softmax score and penalizing object hypotheses that produce ambiguous object mappings. Thus, higher values indicate better object hypotheses that enable clear transforms of few objects. Object hypotheses are ranked by their average maximum softmax value and considered by our solver in descending order.

\begin{table}[t]
    \caption{Our solver's object hypotheses.}
    \begin{center}
        \begin{tabular}{p{0.15\linewidth}p{0.6\linewidth}}
            \toprule
            \multicolumn{1}{c}{Hypothesis} & \multicolumn{1}{c}{Description} \\
            \midrule
            \myoperation{8-Connected} & Each group of 8-connected pixels (i.e., pixels sharing an edge or corner) of the same colour is an object. \\
            \myoperation{4-Connected} & Each group of 4-connected pixels (i.e., pixels sharing an edge) of the same colour is an object. \\
            \myoperation{Vertical} & Each group of vertically contiguous pixels of the same colour is an object. \\
            \myoperation{Horizontal} & Each group of horizontally contiguous pixels of the same colour is an object. \\
            \myoperation{Colour} & All pixels of the same colour, regardless of spatial contiguity, are an object. \\
            \myoperation{Pixel} & Each non-zero (coloured) pixel is its own object. \\
            \bottomrule
        \end{tabular}
    \end{center}
    \label{tab:object_hypotheses}
\end{table}

The second heuristic narrows the search space of object-object correspondences. Instead of considering how each input object can transform into each output object, our solver only considers actions applied to the input object most similar to the output object. Dissimilar objects usually cannot be connected by simple transforms, so this heuristic eliminates nonsensical action hypotheses.

The third heuristic narrows the search space of actions. Instead of considering how each operation can transform the input object into the output object, our solver considers only those operations affecting the least similar property of the objects. Actions need only alter dissimilar object properties, so this heuristic removes useless operations from search.

After this stage (see Algorithm \ref{alg:abduction}), our solver has specific hypotheses about what constitutes a task-relevant object and what actions permit simple transformations in all demonstrations. 

\begin{algorithm}[b]
    \caption{Demonstration Abduction}
    \label{alg:abduction}
    \begin{algorithmic}[1]
        \Require{Demonstrations $\mathcal{D}$}
        \State{$hypotheses \gets$ \Call{ObjectHypothesisHeuristic}{$\mathcal{D}$}} \Comment{Object hypotheses ranked by likelihood}
        \For{\textbf{each} hypothesis $\mathcal{H}$ \textbf{in} $hypotheses$}
            \State{$\mathcal{S} \gets \emptyset$} \Comment{Set of action sets explaining each output object}
            \For{$(\mathbf{G}_{I_d}, \mathbf{G}_{O_d}) \in \mathcal{D}$}
                \State{$\mathcal{O}_{I_d} \gets$ \Call{PerceiveObjects}{$\mathcal{H}$, $\mathbf{G}_{I_d}$}} \Comment{Input objects}
                \State{$\mathcal{O}_{O_d} \gets$ \Call{PerceiveObjects}{$\mathcal{H}$, $\mathbf{G}_{O_d}$}} \Comment{Output objects}
                \For{$O_j \in \mathcal{O}_{O_d}$}
                    \State{$O_i \gets$ \Call{ObjectCorrespondenceHeuristic}{$\mathcal{O}_{I_d}$, $O_j$}} \Comment{Corresponding input object}
                    \State{$operations \gets$ \Call{ObjectActionHeuristic}{$O_i$, $O_j$}} \Comment{Plausible operations for mapping}
                    \State{$\mathcal{A}_k \gets$ \Call{GetPartialActionSet}{$O_i$, $O_j$, $operations$}} \Comment{Action set for output object}
                    \State{$\mathcal{S} \gets \mathcal{S} ~ \cup ~ \{ \mathcal{A}_k \}$}
                \EndFor
            \EndFor
            \State{$\mathcal{A}, cost \gets$ \Call{MinimumHittingSet}{$\mathcal{S}$}} \Comment{Minimal action set and associated cost}
            \If{$\mathcal{A}$ successful \textbf{and} $cost$ acceptable}
                \State{\Return{$\mathcal{H}$, $\mathcal{A}$}}
            \EndIf
        \EndFor
    \end{algorithmic}
\end{algorithm}

\subsubsection{Rule Induction}

Second, our solver induces general object-centric rules explaining the abduced actions. Finding isolated actions explaining each demonstration individually is important, but misses the task's underlying phenomena. Solvers must develop some transferable understanding of the task because ARC-AGI requires generating the correct output grid for arbitrary valid input grids, demonstration or not. In our solver's object-centric view, this means inferring, by both search and learning, what features of the input objects cause them to be transformed in each particular way. In other words, our solver must determine when and, possibly, how each operation should be applied.

Our solver's solution program requires a rule for the conditional application of each operation in the action set. As such, our solver must develop a predictor modelling the condition for whether each operation applies and a predictor for each parameter of each operation. Both types of predictors are implemented as NNs because the high-dimensional distributed object representations produced by the VSA are amenable to neural learning. Although we do not use deep learning, these VSA representations act as deep-learned embeddings; the VSA's inductive biases enable these NNs to better discover the abstractions underlying the conditions and learn the mappings determining the parameters.

Neural learning methods, especially over-parametrized NNs in the sample-few regime, are susceptible to overfitting and shortcut learning. Such NNs tend to get distracted by spurious correlations in the data and often fail to learn the simplest decision criteria. Thus, selecting the input features to our solver's predictor NNs is important. In ARC-AGI tasks, both when and how an operation is applied usually depends on only one or two properties of the input object; the NNs modelling these rules should therefore be exposed only to the subset of relevant properties. Consider ARC-AGI task \texttt{a61f2674} (see Fig. \ref{fig:task_a61f2674}); when and how the \myoperation{Recolour} operation applies depends only on an object's shape property and not its colour or centre properties. Here, supplying the predictor NNs with objects' colour and centre properties serves only to impede learning. Although each predictor NN models System 1, discovering which properties are important for each predictor NN is a search task performed by System 2. Fundamentally, useful properties generalize across demonstrations and useless ones do not.

We believe humans determine which properties are useful by carefully verifying their hypotheses on the demonstrations. Our solver models this process using cross-validation. For each demonstration, our solver, using a particular property hypothesis, trains a predictor NN on data from all other demonstrations and tests the predictor NN on data from the held out demonstration. The aggregate performance across all demonstrations measures the generalizability of the property hypothesis; the highest-scoring hypothesis is selected to learn the final condition. This approach works for many tasks, but not all; some tasks require all demonstrations together to fully understand the rules.

Our solver develops an \textit{operation predictor}, $\mathcal{R}_O$, for each rule modelling the condition for whether the corresponding operation applies to an object. Each operation predictor is an NN mapping the representations of certain properties of an input object onto a probability value signifying whether this operation should be applied to this object. Training input data are the representations of the relevant properties of each input object, and the training labels are binary values indicating whether each input object was subject to this operation. When all labels are positive, our solver assumes the operation applies to all input objects and makes the condition vacuous; an NN is only used when the labels are not uniform.

\begin{figure}
    \centering
    \begin{minipage}{0.47\textwidth}
        \centering
        \includegraphics[width=\linewidth]{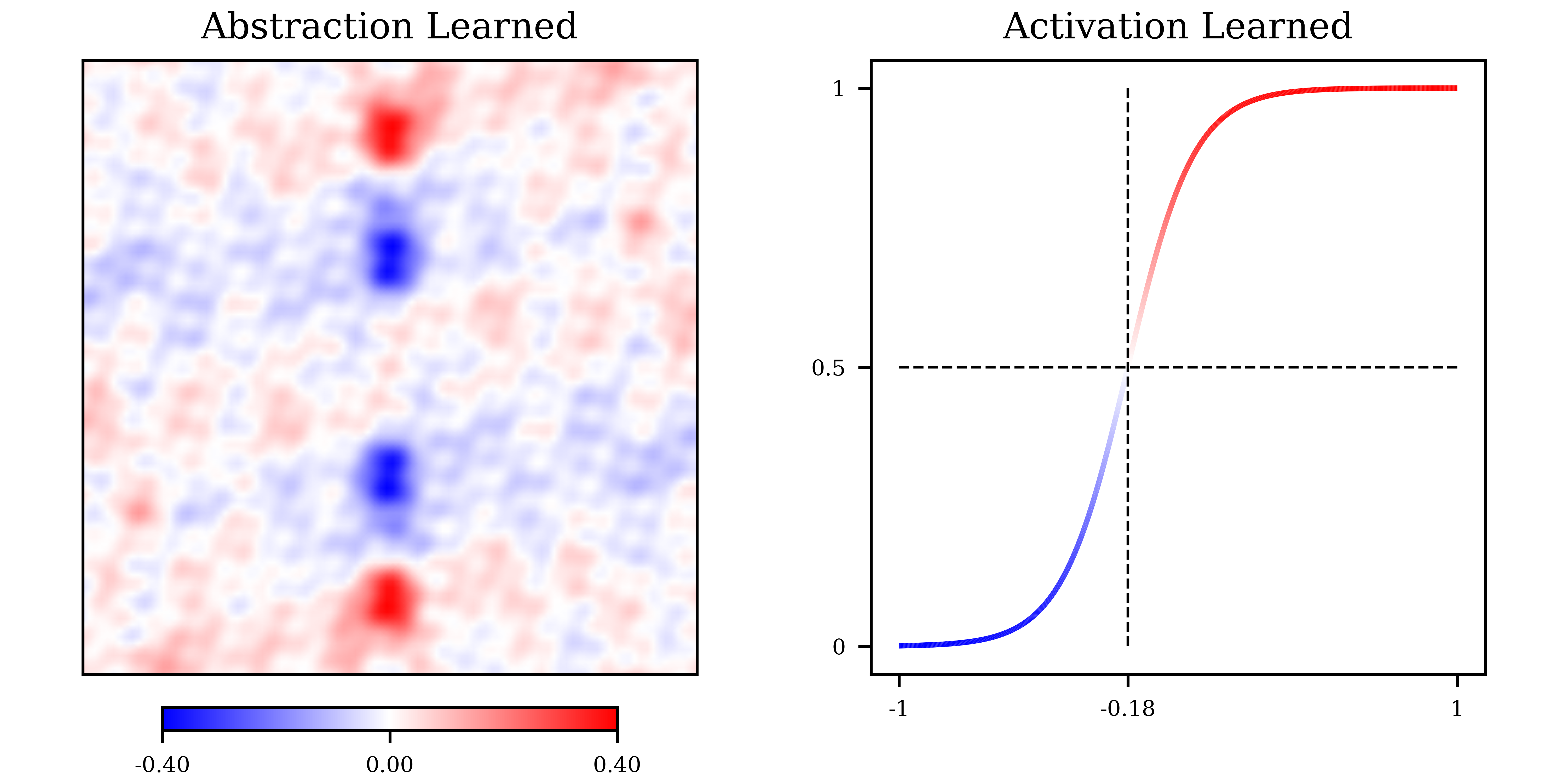}
        \caption{Visualization of the interpretation of the learned \myoperation{Recolour} operation predictor NN for ARC-AGI task \texttt{a61f2674} (see Fig. \protect{\ref{fig:task_a61f2674}}). The abstraction subfigure (left) shows the prototype learned by the NN, determined here by the input object's shape property. The activation subfigure (right) shows the decision nonlinearity learned by the NN. From these, we can verify the NN selects for short and tall objects. Short objects' shape representations exist only in the centre, producing near-zero similarity mapping onto a likely probability. Medium-height objects' shape representations reach the negative similarity regions indicated by blue, producing a negative similarity mapping onto an unlikely probability. Finally, tall objects' shape representations reach the positive similarity regions indicated by red, producing a net positive similarity mapping onto a likely probability. These representations use $N \! = \! 1024$-dimensional vectors.}
        \label{fig:a61f2674_operation_predictor}
    \end{minipage}
    \hfill
    \begin{minipage}{0.47\textwidth}
        \centering
        \includegraphics[width=\linewidth]{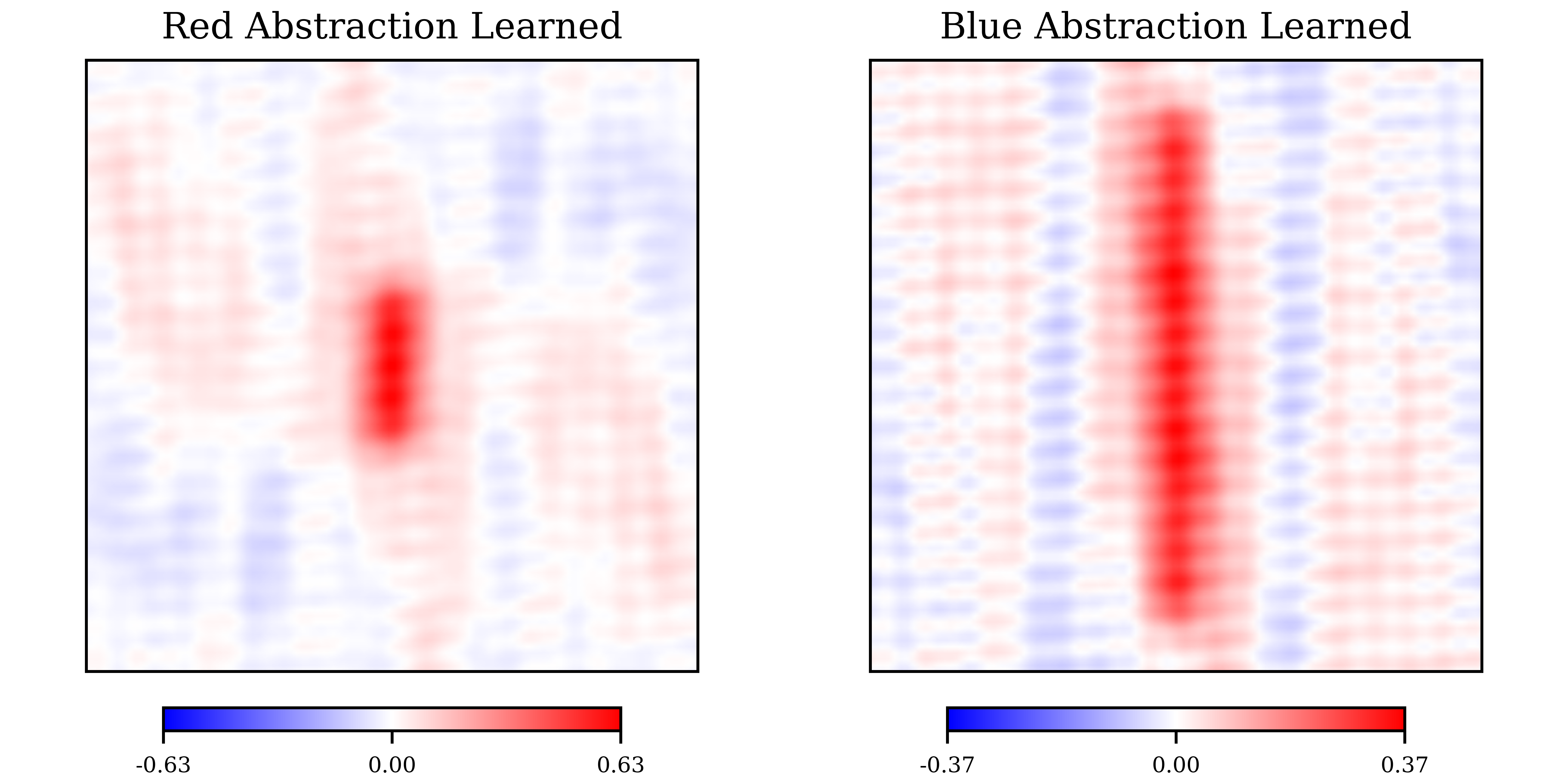}
        \caption{Visualization of the interpretation of the learned \myoperation{Recolour} parameter predictor NN for ARC-AGI task \texttt{a61f2674} (see Fig. \protect{\ref{fig:task_a61f2674}}). The red subfigure (left) shows the prototype for objects recoloured red learned by the NN, determined here by the input object’s shape property. The blue subfigure (right) shows the prototype for objects recoloured blue learned by the NN, again determined here by the input object’s shape property. From these, we can verify the NN predicts red for short objects and blue for tall objects. These representations use $N \! = \! 1024$-dimensional vectors.}
        \label{fig:a61f2674_parameter_predictor}
    \end{minipage}
\end{figure}

The operation predictor is a single-layer NN with a learned nonlinearity trained by stochastic gradient descent (SGD). This means the NN learns a single weight vector to transform the representation of the relevant properties of an input object with via a dot product. Thus, the weight vector, if normalized, is itself a valid VSA vector. As such, we restrict the weight vector to unit length during training and learn parameters controlling the steepness and threshold of the decision nonlinearity. The learned nonlinearity enables the NN to match exact patterns for conditions based on discrete features and to model fuzzy abstractions for conditions based on high-level features. With this construction, the learned weight vector represents a prototype of objects subject to the operation. The NN works by computing the similarity between its learned prototype and the input object's representation and applying a sigmoid nonlinearity mapping this similarity onto a probability, thereby predicting an input object as more likely to be subject to the relevant operation the more similar it is to the prototype (see Fig. \ref{fig:a61f2674_operation_predictor}). To accelerate training, we initialize the NN weight vector with the superposition of the input object representations weighted by outcome. We have experimented with more complex, deeper learning architectures, but this simple approach is the most interpretable and theoretically-founded.

Our solver also develops \textit{parameter predictors}, $\mathcal{R}_P$, for each parameter of each operation. Each parameter predictor is an NN mapping the representations of certain properties of an input object onto a VSA vector representing the value of the operation's input parameter. Training input data are the representations of the relevant properties of each input object, and the training labels are the representations of the abduced parameters used when each input object was subject to this operation. When all labels are the same, our solver assumes this parameter is the same for all input objects and always predicts this parameter. Additionally, when all labels match one of the input objects' properties, our solver assumes this parameter always matches that property of the input object and simply predicts the parameter to be this property. An NN is only used otherwise.

Each parameter predictor is a single-layer NN trained by SGD. This means the NN learns a single square weight matrix performing a linear mapping from one VSA vector onto another VSA vector. A single linear transform, although simple, is usually sufficient; because of the VSA's nonlinear encoding scheme, linear functions in the embedding space can express complex nonlinear functions in the feature space. With this construction, the workings of this NN are less interpretable, but we can sometimes cast its operation as a comparison to prototypes too. Given the learned weight matrix, we can find the optimal input representation to most strongly produce a particular output parameter. In this view, the NN works by memorizing the training mappings and, powered by the VSA representations, constructing a reasonable interpolation of the underlying function (see Fig. \ref{fig:a61f2674_parameter_predictor}). To accelerate training and bias the NN towards VSA-based solutions, we initialize the NN weights with the average circulant matrix performing the binding operation mapping each input vector to its corresponding label. Again, we have experimented with more complex learning architectures, but this simple approach is the most interpretable and theoretically-founded.

The rule induction process involves, for each operation in the action set, iteratively trying to generalize observations into simple rules based on different object properties. This requires intelligently traversing the object property space. As before, we propose a VSA-based heuristic for this task modelling System 1.

This heuristic proposes an intelligent order to traverse the object property space. Instead of trying to learn from random combinations of object properties, our solver first tries to learn from the most relevant object properties. Fundamentally, a relevant object property, meaning one that is likely to explain the condition or parameter mapping, is usually similar among all objects transformed similarly and different between each pair of objects transformed differently. This heuristic computes the average similarity between the properties of all pairs of objects transformed in the same way and the average similarity between the properties of all pairs of objects transformed in different ways, and takes the difference of their squares. We take the softmax of the resulting vector to capture the relative likelihood of each property explaining the application of each operation and the mapping underlying each parameter. Similar objects are likely to be transformed similarly, so first considering those object properties that are most similar avoids unlikely explanations.

After this stage (see Algorithm \ref{alg:induction}), our solver has developed a program as a set of rules describing when and how to apply each operation to input objects.

\begin{algorithm}[t]
    \caption{Rule Induction}
    \label{alg:induction}
    \begin{algorithmic}[1]
        \Require{Input objects $\mathcal{O}_{I}$, Action set $\mathcal{A}$}
        \State{$\mathcal{P} \gets \emptyset$} \Comment{Program comprising a set of rules}
        \For{$operation \in \mathcal{A}$}
            \State{$labels \gets$ \Call{GetObjectOperationLabels}{$\mathcal{A}$, $operation$}} \Comment{Known from abduction stage}
            \State{$properties \gets$ \Call{ObjectPropertyHeuristic}{$\mathcal{O}_{I}$, $labels$}} \Comment{Properties ranked by relevance}
            \State{$\mathcal{R}_O \gets$ \Call{LearnNN}{$\mathcal{O}_{I}$, $labels$, $properties$}} \Comment{Bypassed if all labels are the same}
            \For{\textbf{each} $parameter$ required by $operation$}
                \State{$labels \gets$ \Call{GetObjectParameterLabels}{$\mathcal{A}$, $operation$, $parameter$}}
                \State{$properties \gets$ \Call{ObjectPropertyHeuristic}{$\mathcal{O}_{I}$, $labels$}}
                \State{$\mathcal{R}_P \gets$ \Call{LearnNN}{$\mathcal{O}_{I}$, $labels$, $properties$}} \Comment{Bypassed if all labels are the same or unchanged}
            \EndFor
            \State{$\mathcal{P} \gets \mathcal{P} ~ \cup ~ \{ (operation, \mathcal{R}_O, \mathcal{R}_P) \}$}
        \EndFor
        \State{\Return{$\mathcal{P}$}}
    \end{algorithmic}
\end{algorithm}

\subsubsection{Answer Deduction}

Third, our solver deduces the output grid for each query. Any solver must apply whatever understanding it has obtained about the demonstrations to the queries. This requires generating an output grid from scratch.

By our solution structure, this means sizing the output grid and predicting and executing the actions for each input object. Each input object is considered separately by each rule; if the object satisfies the rule's learned condition, then the rule's action, with parameters predicted as needed, is executed and the new object is added to the output grid. Objects may be subject to multiple rules, producing multiple output objects, and rules may be applied to multiple objects.

After this stage (see Algorithm \ref{alg:deduction}), our solver has solved the task.

\begin{algorithm}[b]
    \caption{Answer Deduction}
    \label{alg:deduction}
    \begin{algorithmic}[1]
        \Require{Object hypothesis $\mathcal{H}$, Program $\mathcal{P}$, Queries $\mathcal{Q}$}
        \State{$results \gets [ ~ ]$} \Comment{Test output grids}
        \For{$\mathbf{G}_{I_q} \in \mathcal{Q}$}
            \State{$\mathcal{O}_{I_q} \gets$ \Call{PerceiveObjects}{$\mathcal{H}$, $\mathbf{G}_{I_q}$}} \Comment{Input objects}
            \State{$\mathcal{O}_{O_q} \gets \emptyset$} \Comment{Output objects}
            \For{$O_i \in \mathcal{O}_{I_q}$}
                \For{$( operation, \mathcal{R}_O , \mathcal{R}_P ) \in \mathcal{P}$}
                    \If{$\mathcal{R}_O ( O_i ) \geq 50\%$} \Comment{Threshold is arbitrarily 50\%}
                        \State{$\mathcal{O}_{O_q} \gets \mathcal{O}_{O_q} ~ \cup ~$ \Call{ExecuteAction}{$O_i$, $operation$, $\mathcal{R}_P ( O_i )$}}
                    \EndIf
                \EndFor
            \EndFor
            \State{$results \gets results ~ \cup ~ \mathcal{O}_{O_q}$}
        \EndFor
        \State{\Return{$results$}}
    \end{algorithmic}
\end{algorithm}

\section{Results}
\label{sec:results}

We evaluate the performance of our solver on ARC-AGI and related benchmarks. Because our solver generates task solutions as neurosymbolic programs, we can probe performance by executing the solution program on both the training input grids and the testing input grids. When our solver can correctly solve the demonstrations, it has some understanding of the task; when our solver can correctly solve the queries, it has a true, generalizable understanding of the task. Therefore, demonstration performance indicates whether our solver can conceptualize the objects and actions underlying task transformations, but query performance reveals whether our solver can extract and apply the underlying patterns. We report both \textit{accuracy}, the proportion of all demonstrations or queries in the dataset solved correctly, and \textit{task accuracy}, the proportion of tasks for which all associated demonstrations or queries were solved correctly. The main performance metric, except where noted, is query task accuracy. All experiments use representations with $N \! = \! 4096$-dimensional vectors.

\subsection{Main Benchmarks}

Our solver scores $10.8\%$ on ARC-AGI-1-Train and $3.0\%$ on ARC-AGI-1-Eval (see Table \ref{tab:arc_results}). We report performance on the training splits in addition to the evaluation splits because they are more approachable (note that the training splits are not used for training; our solver does not require any pre-training). As expected, our solver performs better on ARC-AGI-1 than on ARC-AGI-2 and performs better on training splits than evaluation splits.

\begin{table}
    \caption{Our solver's performance on ARC-AGI. We report performance on each benchmark split separately.}
    \begin{center}
        \begin{tabular}{lcccc}
            \toprule
            \multicolumn{1}{c}{\multirow{2}{*}{Benchmark Split}} & \multicolumn{2}{c}{Demonstrations} & \multicolumn{2}{c}{Queries} \\
            \cmidrule{2-5} & Acc. ($\%$) & Task Acc. ($\%$) & Acc. ($\%$) & Task Acc. ($\%$) \\
            \midrule
            ARC-AGI-1-Train ($n=400$) & $57.5$ & $48.8$ & $12.7$ & $10.8$ \\
            ARC-AGI-1-Eval ($n=400$) & $43.6$ & $33.5$ & $3.3$ & $3.0$ \\
            ARC-AGI-2-Train ($n=1000$) & $46.7$ & $35.9$ & $6.5$ & $5.8$ \\
            ARC-AGI-2-Eval ($n=120$) & $19.5$ & $11.7$ & $0.0$ & $0.0$ \\
            \bottomrule
        \end{tabular}
    \end{center}
    \label{tab:arc_results}
\end{table}

\subsection{Related Benchmarks}

Because of ARC-AGI's extreme difficulty, many simpler versions capturing certain important features of the benchmark have been made. We also evaluate the performance of our solver on some of these benchmarks.

\subsubsection{Sort-of-ARC}

Sort-of-ARC \citep{sortofarc} is a collection of 1000 restricted ARC-AGI-like tasks. Each task comprises $|\mathcal{D}| = 5$ demonstrations and $|\mathcal{Q}| = 1$ query; all grids are $20 \times 20$ and contain three non-overlapping objects of random colour, location, and $3 \times 3$ shape. The solution is always to move all objects matching a randomly-chosen colour or shape by one pixel in one of the four cardinal directions and leave all other objects unchanged. Although simple, Sort-of-ARC directly tests condition-action learning.

Our solver scores $94.5\%$ on Sort-of-ARC (see Table \ref{tab:sortofarc_results}). Notably, our solver performs better on tasks with a colour-based condition than those with a shape-based condition. This is because different colour representations have near-zero similarity but different shape representations may have very high similarity. As a result, without diverse negative examples in the demonstrations, the operation predictors tend to over-generalize shape abstractions (e.g., our solver may move U-shaped objects when the solution is to move all O-shaped objects and the demonstrations lack U-shaped or similarly-shaped objects). This generalization capacity enables our solver to solve more complex tasks, but sometimes poses challenges for simple tasks requiring strict rules.

\begin{table}[b]
    \caption{Our solver's performance on Sort-of-ARC \protect{\citep{sortofarc}}. We report performance on tasks with a colour-based condition and tasks with a shape-based condition separately. Our version of the dataset was randomly generated according to \protect{\citeauthor{sortofarc}}'s \protect{\citeyearpar{sortofarc}} description.}
    \begin{center}
        \begin{tabular}{lcccc}
            \toprule
            \multicolumn{1}{c}{\multirow{2}{*}{Benchmark Split}} & \multicolumn{2}{c}{Demonstrations} & \multicolumn{2}{c}{Queries} \\
            \cmidrule{2-5} & Acc. ($\%$) & Task Acc. ($\%$) & Acc. ($\%$) & Task Acc. ($\%$) \\
            \midrule
            All ($n=1000$) & $99.7$ & $99.4$ & --- & $94.5$ \\
            ~~~~ Colour ($n=500$) & $100.0$ & $100.0$ & --- & $100.0$ \\
            ~~~~ Shape ($n=500$) & $99.4$ & $98.8$ & --- & $89.0$ \\
            \bottomrule
        \end{tabular}
    \end{center}
    \label{tab:sortofarc_results}
\end{table}

\subsubsection{1D-ARC}

1D-ARC \citep{1darc} is a collection of 900 one-dimensional ARC-AGI-like tasks. Each task comprises $|\mathcal{D}| = 3$ demonstrations and $|\mathcal{Q}| = 1$ query; all grids are a single row of pixels. Tasks are divided into 18 types, such as ``Move Dynamic'' and ``Recolour By Size'', each capturing some important operation in ARC-AGI. Each type contains 50 random instantiations of the operation. Designed to probe how LLMs solve ARC-AGI, 1D-ARC thoroughly tests the application of particular operations.

Our solver scores $83.1\%$ on 1D-ARC (see Table \ref{tab:1darc_results}). Our solver can consistently solve tasks involving operations within its DSL, such as those requiring recolouring (e.g., Recolour by Odd Even and Recolour by Size), recentring (e.g., Move 1, Move Dynamic, and Scaling), and reshaping (e.g., Fill, Hollow, and Pattern Copy). But, our solver falters on tasks requiring operations outside of its DSL, such as reflecting (e.g., Flip and Mirror). Still, our solver vastly outperforms GPT-4 \citep{gpt4} on this benchmark at a tiny fraction of the computational cost.

\begin{table}
    \caption{Our solver's performance on 1D-ARC \protect{\citep{1darc}}. We report performance on each task type separately. We compare our results to results obtained by directly prompting GPT-4 \protect{\citep{gpt4}}, as reported in the original work \protect{\citep{1darc}}.}
    \begin{center}
        \begin{tabular}{lccccc}
            \toprule
            \multicolumn{1}{c}{\multirow{2}{*}{Benchmark Split}} & \multicolumn{2}{c}{Demonstrations} & \multicolumn{2}{c}{Queries} & \multicolumn{1}{c}{\multirow{2}{*}{GPT-4}} \\
            \cmidrule{2-5} & Acc. ($\%$) & Task Acc. ($\%$) & Acc. ($\%$) & Task Acc. ($\%$) \\
            \midrule
            All ($n=900$) & $97.1$ & $92.9$ & --- & $83.1$ & --- \\
            ~~~~ Move 1 ($n=50$) & $100.0$ & $100.0$ & --- & $\mathbf{100.0}$ & $66.0$ \\
            ~~~~ Move 2 ($n=50$) & $100.0$ & $100.0$ & --- & $\mathbf{100.0}$ & $26.0$ \\
            ~~~~ Move 3 ($n=50$) & $100.0$ & $100.0$ & --- & $\mathbf{100.0}$ & $24.0$ \\
            ~~~~ Move Dynamic ($n=50$) & $100.0$ & $100.0$ & --- & $\mathbf{90.0}$ & $22.0$ \\
            ~~~~ Move 2 Towards ($n=50$) & $100.0$ & $100.0$ & --- & $\mathbf{98.0}$ & $34.0$ \\
            ~~~~ Fill ($n=50$) & $100.0$ & $100.0$ & --- & $\mathbf{100.0}$ & $66.0$ \\
            ~~~~ Padded Fill ($n=50$) & $100.0$ & $100.0$ & --- & $\mathbf{46.0}$ & $26.0$ \\
            ~~~~ Hollow ($n=50$) & $100.0$ & $100.0$ & --- & $\mathbf{100.0}$ & $56.0$ \\
            ~~~~ Flip ($n=50$) & $77.3$ & $44.0$ & --- & $12.0$ & $\mathbf{70.0}$ \\
            ~~~~ Mirror ($n=50$) & $94.0$ & $90.0$ & --- & $\mathbf{28.0}$ & $20.0$ \\
            ~~~~ Denoise ($n=50$) & $100.0$ & $100.0$ & --- & $\mathbf{100.0}$ & $36.0$ \\
            ~~~~ Denoise Multicolour ($n=50$) & $100.0$ & $100.0$ & --- & $\mathbf{98.0}$ & $60.0$ \\
            ~~~~ Pattern Copy ($n=50$) & $100.0$ & $100.0$ & --- & $\mathbf{100.0}$ & $36.0$ \\
            ~~~~ Pattern Copy Multicolour ($n=50$) & $100.0$ & $100.0$ & --- & $\mathbf{100.0}$  & $38.0$ \\
            ~~~~ Recolour by Odd Even ($n=50$) & $100.0$ & $100.0$ & --- & $\mathbf{96.0}$ & $32.0$ \\
            ~~~~ Recolour by Size ($n=50$) & $100.0$ & $100.0$ & --- & $\mathbf{100.0}$ & $28.0$ \\
            ~~~~ Recolour by Size Comparison ($n=50$) & $76.0$ & $38.0$ & --- & $\mathbf{32.0}$ & $20.0$ \\
            ~~~~ Scaling ($n=50$) & $100.0$ & $100.0$ & --- & $\mathbf{96.0}$ & $88.0$ \\
            \bottomrule
        \end{tabular}
    \end{center}
    \label{tab:1darc_results}
\end{table}

\subsubsection{KidsARC}

KidsARC \citep{kidsarc} is a collection of 17 single-demonstration ARC-AGI-like tasks. Each task comprises $|\mathcal{D}| = 1$ demonstration and $|\mathcal{Q}| = 1$ query; all grids in the 9 simple tasks are $3 \times 3$, and all grids in the 8 concept tasks are $5 \times 5$. The 9 simple tasks invoke elementary concepts and can be presented as $A:B::C:D$ or as $A:C::B:D$; the 8 concept tasks invoke higher-level abstractions. Designed to understand how children approach ARC-AGI, KidsARC uniquely tests extreme generalization.

Our solver scores $57.7\%$ on KidsARC (see Table \ref{tab:kidsarc_results}). Our solver can abduce the correct operations (e.g., recolouring, recentring, and reshaping) and even learn useful abstractions from a single demonstration (e.g., objects near the top of the grid and dominant versus noise objects). But, our solver falters when extreme generalization of a high-level concept is needed from a single example (e.g., objects between other objects and objects with the majority colour). Again, our solver outperforms many LLMs on this benchmark, still at a tiny fraction of the computational cost.

\begin{table}[b]
    \caption{Our solver's performance on KidsARC \protect{\citep{kidsarc}}. We report performance on simple and concept tasks separately. We compare our results to aggregate results obtained by directly prompting various LLMs, including GPT-4 \protect{\citep{gpt4}}, as reported in the original work \protect{\citep{kidsarc}}.}
    \begin{center}
        \begin{tabular}{lccccc}
            \toprule
            \multicolumn{1}{c}{\multirow{2}{*}{Benchmark Split}} & \multicolumn{2}{c}{Demonstrations} & \multicolumn{2}{c}{Queries} & \multicolumn{1}{c}{\multirow{2}{*}{LLMs}} \\
            \cmidrule{2-5} & Acc. ($\%$) & Task Acc. ($\%$) & Acc. ($\%$) & Task Acc. ($\%$) \\
            \midrule
            All ($n=26$) & --- & $92.3$ & --- & $57.7$ & --- \\
            ~~~~ Simple ($n=18$) & --- & $94.4$ & --- & $\mathbf{66.7}$ & $33.2$ \\
            ~~~~ Concept ($n=8$) & --- & $87.5$ & --- & $\mathbf{37.5}$ & $11.9$ \\
            \bottomrule
        \end{tabular}
    \end{center}
    \label{tab:kidsarc_results}
\end{table}

\subsubsection{ConceptARC}

ConceptARC \citep{conceptarc} is a collection of 176 complex ARC-AGI-like tasks. Each task comprises a variable number of demonstrations and $|\mathcal{Q}| = 3$ queries; like ARC-AGI, grid sizes are variable. Tasks are divided into 16 concept groups, such as ``Above and Below'' and ``Same and Different'', each capturing some important abstraction in ARC-AGI. Each group contains a minimal example alongside 10 carefully-designed instantiations of the concept. Designed to probe how completely ARC-AGI solvers understand the concepts they use, ConceptARC thoroughly tests the comprehension of particular concepts.

Our solver scores $20.5\%$ on ConceptARC (see Table \ref{tab:conceptarc_results}). Our solver performs well on spatial concepts (e.g., Above and Below, Filled and Not Filled, Horizontal and Vertical, and Top and Bottom 2D) because of the inductive biases provided by SSP representations. But, our solver falters on tasks requiring interpolative object perception (e.g., Clean Up, Complete Shape, Copy, and Top and Bottom 3D) and advanced explicit reasoning (e.g., Count, Order, and Same and Different). Still, our solver performs comparably to GPT-4 \citep{gpt4} on this benchmark, again at a tiny fraction of the computational cost.

\begin{table}
    \caption{Our solver's performance on ConceptARC \protect{\citep{conceptarc}}. We report performance on the minimal tasks as well as each task type separately. We compare our results to results obtained by directly prompting GPT-4 \protect{\citep{gpt4}}, as reported in the original work \protect{\citep{conceptarc}}. Note that ConceptARC measures performance as the proportion of queries solved correctly, not the proportion of tasks with all queries solved correctly.}
    \begin{center}
        \begin{tabular}{lccccc}
            \toprule
            \multicolumn{1}{c}{\multirow{2}{*}{Benchmark Split}} & \multicolumn{2}{c}{Demonstrations} & \multicolumn{2}{c}{Queries} & \multicolumn{1}{c}{\multirow{2}{*}{GPT-4}} \\
            \cmidrule{2-5} & Acc. ($\%$) & Task Acc. ($\%$) & Acc. ($\%$) & Task Acc. ($\%$) \\
            \midrule
            All ($n=176$) & $79.1$ & $73.3$ & $20.5$ & $11.4$ & --- \\
            ~~~~ Minimal ($n=16$) & $91.3$ & $87.5$ & $52.1$ & $31.2$ & --- \\
            ~~~~ Above and Below ($n=10$) & $87.5$ & $90.0$ & $\mathbf{26.7}$ & $20.0$ & $23.3$ \\
            ~~~~ Center ($n=10$) & $80.0$ & $70.0$ & $23.3$ & $10.0$ & $\mathbf{33.3}$ \\
            ~~~~ Clean Up ($n=10$) & $91.3$ & $90.0$ & $6.7$ & $0.0$ & $\mathbf{20.0}$ \\
            ~~~~ Complete Shape ($n=10$) & $90.5$ & $90.0$ & $3.3$ & $0.0$ & $\mathbf{23.3}$ \\
            ~~~~ Copy ($n=10$) & $65.2$ & $70.0$ & $0.0$ & $0.0$ & $\mathbf{23.3}$ \\
            ~~~~ Count ($n=10$) & $11.1$ & $10.0$ & $3.3$ & $0.0$ & $\mathbf{13.3}$ \\
            ~~~~ Extend To Boundary ($n=10$) & $93.1$ & $90.0$ & $3.3$ & $0.0$ & $\mathbf{6.7}$ \\
            ~~~~ Extract Objects ($n=10$) & $43.5$ & $40.0$ & $\mathbf{20.0}$ & $20.0$ & $3.3$ \\
            ~~~~ Filled and Not Filled ($n=10$) & $82.8$ & $70.0$ & $\mathbf{20.0}$ & $0.0$ & $16.7$ \\
            ~~~~ Horizontal and Vertical ($n=10$) & $96.0$ & $90.0$ & $\mathbf{36.7}$ & $30.0$ & $26.7$ \\
            ~~~~ Inside and Outside ($n=10$) & $69.0$ & $60.0$ & $\mathbf{20.0}$ & $10.0$ & $10.0$ \\
            ~~~~ Move To Boundary ($n=10$) & $92.0$ & $90.0$ & $\mathbf{30.0}$ & $10.0$ & $20.0$ \\
            ~~~~ Order ($n=10$) & $76.2$ & $70.0$ & $10.0$ & $0.0$ & $\mathbf{26.7}$ \\
            ~~~~ Same and Different ($n=10$) & $84.8$ & $70.0$ & $6.7$ & $0.0$ & $\mathbf{16.7}$ \\
            ~~~~ Top and Bottom 2D ($n=10$) & $94.1$ & $80.0$ & $\mathbf{60.0}$ & $50.0$ & $23.3$ \\
            ~~~~ Top and Bottom 3D ($n=10$) & $80.6$ & $70.0$ & $6.7$ & $0.0$ & $\mathbf{20.0}$ \\
            \bottomrule
        \end{tabular}
    \end{center}
    \label{tab:conceptarc_results}
\end{table}

\subsubsection{MiniARC}

MiniARC \citep{miniarc} is a collection of 149 miniature ARC-AGI-like tasks. Each task comprises a variable number of demonstrations and $|\mathcal{Q}| = 1$ query; all grids are $5 \times 5$. Like ARC-AGI, task content is unconstrained. Designed to simplify ARC-AGI without compromising its core tenets, MiniARC tests reasoning without grid size effects.

Our solver scores $13.4\%$ on MiniARC (see Table \ref{tab:miniarc_results}). Notably, our solver can solve multiple tasks requiring conceptualizing and implementing arbitrary object mappings (e.g., tasks \texttt{l6aftgp22mlspnm1zxb}, \texttt{l6acypud56b4m1z5409}, and \texttt{l6afcchh0wfew0rhswzq}).

\begin{table}[b]
    \caption{Our solver's performance on MiniARC \protect{\citep{miniarc}}.}
    \begin{center}
        \begin{tabular}{lcccc}
            \toprule
            \multicolumn{1}{c}{\multirow{2}{*}{Benchmark Split}} & \multicolumn{2}{c}{Demonstrations} & \multicolumn{2}{c}{Queries} \\
            \cmidrule{2-5} & Acc. ($\%$) & Task Acc. ($\%$) & Acc. ($\%$) & Task Acc. ($\%$) \\
            \midrule
            All ($n=149$) & $69.9$ & $55.0$ & --- & $13.4$ \\
            \bottomrule
        \end{tabular}
    \end{center}
    \label{tab:miniarc_results}
\end{table}

\section{Discussion}
\label{sec:discussion}

Our solver structures solutions as a specific definition of task-relevant objects and a specific program comprising a set of rules describing how objects behave. Objects are represented by high-dimensional vectors encoding their colour, centre, and shape. Program rules are if-then statements that conditionally execute standard actions on certain objects based on object features. This paradigm is flexible enough to express natural solutions to many ARC-AGI tasks, but simple enough for synthesis to remain tractable.

Our solver generates solutions with a heuristic-guided search process and a neurosymbolic learning architecture. First, our solver abduces the simplest objects and actions explaining the demonstrations. Second, our solver generalizes its observations into rules applicable to any valid input grid. Third, our solver applies these rules to answer all queries. This process combines the strengths of symbolic and connectionist AI, leveraging VSA representations for robustness and interpretability in search and learning.

\subsection{Strengths}

Any model of human intelligence must be efficient and interpretable. For ARC-AGI, this is true of both the solution synthesis process and the synthesized solution itself.

Human intelligence is efficient. The brain's power consumption is much lower than that of modern computers, and humans effortlessly solve ARC-AGI. Because our solver uses neither brute force search nor large NNs, its computational demands are relatively small. Our solver considers few solution candidates in depth because simple and cheap heuristics are used to prune the search space, minimizing wasted energy on unnecessary reasoning. Overall, our solver is efficient.

Human intelligence is interpretable. When prompted, humans can detail both the problem solving steps they took and the solution they arrived at. Because intermediate hypotheses are few, all steps taken by our solver are traceable and intentional. The solution programs generated are compact and integrate directly interpretable symbolic expressions with interpretable NNs implementing simple operations on structured data. Overall, our solver is interpretable.

Our methods are applicable to other problems beyond ARC-AGI. We use general VSA representations capable of implementing many features of cognition. For example, Spaun \citep{spaun2,spaun}, the world's largest functional brain model, uses similar representations to closely model several different parts of the brain and perform a wide array of cognitive tasks. Additionally, we apply general principles learned from human psychology to motivate and structure our solver. For example, we explicitly model the interplay of System 1 intuitions and System 2 reasoning \citep{kahneman} in a unified substrate. Our particular implementation and architecture is ARC-AGI-specific, but the underlying principles and tools are general; similar VSA-based approaches may be taken to solve other problems requiring System 1, System 2, or both. Fundamentally, VSA-powered neurosymbolic models represent a universal approach to cognitive modelling and AI.

Our approach furthers progress on ARC-AGI. Chollet, the creator of ARC-AGI, has argued the proper solution to ARC-AGI may be neurosymbolic \citep{arcprize2024}, involving both symbolic program synthesis and connectionist deep learning. VSAs offer a new means of bridging symbolic and connectionist AI for ARC-AGI. To the best of our knowledge, our solver is the first to apply VSAs to ARC-AGI, making our approach unique and novel. Additionally, because we use cognitively and biologically plausible representations in a psychologically-motivated framework free from brute force search and opaque deep NNs, we believe our solver is the most cognitively plausible one yet.

Our solver owes its success to the same strengths of VSAs that make them so effective for other cognitive tasks. First, VSAs bridge symbolicism and connectionism, combining the benefits of both. In a single unified framework, VSAs enable both heuristic-guided search with explicit reasoning and sample-efficient neural learning. Second, VSAs capture many inductive biases humans have. In this sense, VSAs effectively model human intelligence, which remains the best ARC-AGI solver. The required numbers and geometry core knowledge priors are implicitly provided by SSP representations, and objectness is explicitly modelled by our solver in other ways (we leave goal-directedness for future work).

\subsection{Limitations}

Our solver is primarily limited by its inability to generate new grid size change, object, and action hypotheses on-the-fly. The demonstration abduction stage is only capable of applying existing notions of what grid sizes, objects, and actions may be taken from finite sets. This is permissible with a sufficiently-sized hypothesis set because all ARC-AGI tasks must be derived from the core knowledge priors, but our solver's sets are currently not large enough. Furthermore, we do not address the fundamental problem of how these conceptualizations came to be; instead, we assume they have already been acquired. Maintaining cognitive plausibility, we model the process by which humans recombine existing knowledge, represented as inductive biases built into the VSAs and symbolic programs, to solve new tasks. In this sense, our solver is not artificial \textit{general} intelligence because it uses a domain-\textit{specific} language; a true AGI solution must be DSL-open, discovering and applying novel transformations on-the-fly.

Other limitations are more pragmatic. First, our solver cannot apply chained operations to a single persistent input object. This simplifies solution synthesis by narrowing the search space of object-to-object transforms at the cost of preventing arbitrary compositionality of operations, but the \myoperation{Generate} operation can compose \myoperation{Recolour}, \myoperation{Recentre}, and \myoperation{Shape} operations. Second, our solver cannot apply multiple instantiations of one operation to the same input object. This simplifies program representation by restricting the program to one rule for each operation at the cost of solving certain tasks. Third, our solver can conceptualize neither many-to-one nor many-to-many object mappings. This simplifies solution synthesis also by narrowing the search space of object-to-object transforms at the cost of solving tasks requiring comparison between objects, but creative object hypotheses can often circumvent this. Fourth, our solver cannot conceptualize multi-coloured objects. This simplifies object representation by allowing separate colour, centre, and shape properties at the cost of solving certain tasks, but creative actions can often circumvent this. Fifth, our solver cannot conceptualize conditions of high-level object properties. This simplifies object representation and maintains cognitive plausibility at the cost of solving certain tasks, but these properties, such as exact size and symmetry, can often be well-approximated by other object properties, such as shape. All of these limitations are the subject of ongoing work.

\section{Conclusion}
\label{sec:conclusion}

We presented a novel, neurosymbolic, cognitively plausible ARC-AGI solver. Our solver works by VSA-enabled object-centric program synthesis. Grids are represented as collections of objects encoded with a VSA, and solutions are represented as programs comprising a set of rules implemented with a VSA. Solution synthesis uses VSAs to bridge symbolic and connectionist AI; modelling human intelligence, VSA-powered System-1-style heuristics guide a System-2-style intentional search process incorporating abductive, inductive, and deductive reasoning. VSA representations enable sample-efficient neural learning of rules, integrating automated abstraction discovery into the reasoning process. The solution generation process is both efficient and interpretable, maintaining cognitive plausibility. Our solver shows some initial success, scoring $10.8\%$ on ARC-AGI-1-Train and $3.0\%$ on ARC-AGI-1-Eval. Although these results are not state-of-the-art, they indicate VSAs hold promise for ARC-AGI and warrant further investigation. Importantly, our approach is unique; we believe we are the first to apply VSAs to ARC-AGI and, in doing so, have developed the most cognitively plausible ARC-AGI solver yet.

\section*{Acknowledgements}

The authors would like to thank Varun Dhanraj and Marvin Pafla for discussions that helped improve this paper. This work was supported by CFI (52479-10006) and OIT (35768) infrastructure funding as well as the Canada Research Chairs program, NSERC Discovery grant 261453, AFOSR grant FA9550-17-1-0644 and NSERC CGS-M graduate funding.

\end{document}